\RequirePackage{fix-cm}
\documentclass[twocolumn]{svjour3}          
\smartqed  
\usepackage{cite}
\usepackage{amsmath}
\usepackage{amsfonts}
\usepackage{amssymb}
\usepackage{algorithmic}
\usepackage{graphicx}
\usepackage{mathptmx}      
\usepackage{hyperref}
\usepackage{xurl}

\usepackage{pgfplots}
\usepackage{pdfpages}
\usepackage{multirow}
\usepackage{todonotes}
\usepackage{color, colortbl} 
\renewcommand\footnotemark{}

\definecolor{Gray}{gray}{0.95}

\journalname{International Journal of Social Robotics}

\begin{document}
\sloppy

\title{The Child Factor in Child-Robot Interaction



\thanks{}
}

\subtitle{Discovering the Impact of Developmental Stage and Individual Characteristics}

\author{Irina Rudenko \and
        Andrey Rudenko \and 
       Achim J. Lilienthal \and
       Kai O. Arras \and
       Barbara Bruno
}


\institute{Irina Rudenko \at
Child development researcher, Freiburg im Breisgau, Germany
\\
              \email{roudenko.irina@gmail.com}           
            \and
Andrey Rudenko \at
Robert Bosch GmbH, Corporate Research, Stuttgart, Germany
\\
              \email{andrey.rudenko@de.bosch.com}           
            \and
Achim J. Lilienthal \at
Chair Perception for Intelligent Systems (PercInS),
Munich Institute of Robotics and Machine Intelligence (MIRMI),
School of Computation, Information and Technology (CIT),
Technical University of Munich (TUM), Germany, 
and
Mobile Robotics and Olfaction Lab (MRO),
Centre for Applied Autonomous Sensor Systems (AASS), 
Örebro University, Sweden
\\
              \email{achim.j.lilienthal@tum.de}           
            \and
Kai O. Arras \at
Faculty of Computer Science, Electrical Engineering and Information Technology,
University of Stuttgart, Germany
\\
              \email{kai.arras@f05.uni-stuttgart.de}           
            \and
Barbara Bruno \at
Institute for Anthropomatics and Robotics - Socially Assistive Robotics with Artificial Intelligence Lab (IAR-SARAI),
Karlsruhe Institute of Technology (KIT), Germany
\\
              \email{barbara.bruno@kit.edu}           
}

\date{Received: date / Accepted: date}

\maketitle

\begin{abstract}
Social robots, owing to their embodied physical presence in human spaces and the ability to directly interact with the users and their environment, have a great potential to support children in various activities in education, healthcare and daily life.
Child-Robot Interaction (CRI), as any domain involving children, inevitably faces the major challenge of designing generalized strategies to work with unique, turbulent and very diverse individuals.
Addressing this challenging endeavor requires to combine the standpoint of the robot-centered perspective, i.e. what robots technically can and are best positioned to do, with that of the child-centered perspective, i.e. what children may gain from the robot and how the robot should act to best support them in reaching the goals of the interaction.
This article aims to help researchers bridge the two perspectives and proposes to address the development of CRI scenarios with insights from child psychology and child development theories.
To that end, we review the outcomes of the CRI studies, outline common trends and challenges, and identify two key factors from child psychology that impact child-robot interactions, especially in a long-term perspective: developmental stage and individual characteristics. For both of them we discuss prospective experiment designs which support building naturally engaging and sustainable interactions.

\end{abstract}

\section{Introduction}
 
Child-Robot Interaction (CRI), among other Human-Robot Interaction (HRI) topics, is a research area of ever-growing importance and interest, fuelled by the advances in artificial intelligence and robotics, and a worldwide rise of technology use by children \cite{gottschalk2019impacts}. 
The key trigger to researching and developing CRI scenarios, as well as the main reason for their complexity, is the breadth of tasks and challenges that children face during the relatively short and dynamic childhood period.
In less than two decades a child has to master a vast range of skills and obtain a wealth of knowledge in order to become a successful adult. Investigating the feasibility and effectiveness of robot interventions in supporting child development during different childhood periods and activities is thus of paramount importance for our societies.


Quite interestingly, social robots appear to fit naturally in a child’s life, especially in the early years. Thanks to the unique character of children’s mentality, perception and cognition (anthropomorphism, irrationality, animacy, etc. \cite{piaget2003psychology}),
their interactions with robots likely become personal and social, similar to their relationships with pets, toys and some other objects.
This ``natural alignment'' between social robots and children's interests is also highlighted by various recent reviews discussing the potential of Socially Assistive Robots (SAR) in education, healthcare, therapy and as home assistants, and their advantages over virtual agents \cite{logan2019social, kose2015effect,hsiao2015irobiq,hyun2008comparative,jost2012ethological}.
The same reviews, however, also highlight how this potential has yet to be proven in reality.

One of the biggest challenges of CRI research is its intrinsically multi-disciplinary nature \cite{michaud2005autonomous,bartneck2009measurement,baxter2016characterising}, as it sets human-oriented goals (therapeutic, educational, developmental and others) and aims to leverage and engineer robot functionalities to benefit the achievement of such goals.
Balancing the human-centered and robot-centered natures of CRI is a difficult feat to put in practice: state-of-the-art CRI works prioritizing a robot-centered perspective focus on improving the robot itself and validate its proficiency in ad-hoc designed user studies, which may dilute or even miss the end goals of the interaction and overlook working solutions
for child development support \cite{michaud2005autonomous}. Similarly, CRI works prioritizing a human-centered perspective focus on reaching human-oriented goals, possibly neglecting the unique opportunities offered by the robot and often reducing it to a mere puppet or an expensive tablet case \cite{johal2020research}.


This work attempts to support the bridging of the two perspectives, by combining expertise in social robotics and child development to review the state of the art in CRI literature under the two lenses and discuss (i) the goals that a child-robot interaction sets, (ii) the factors that influence the outcome of the interaction, and (iii) how to develop a successful and sustainable long-term CRI scenario.

To this end, in Section \ref{sec:CRI-domains} we survey the state of the art in CRI,  with a particular focus on the  application areas of education and healthcare, which are the most popular and advanced contexts for CRI.
We discuss the outcomes of the studies to outline common trends and challenges in Section \ref{sec:robot-for-a-long-term-interaction} and, in Section \ref{sec:robot-for-the-child}, identify two key factors from child psychology that impact the child-robot interactions in a long-term perspective: developmental stage and individual characteristics.
In Section \ref{sec:age} we thus review the literature in CRI through the child development theory perspective considering the dominant activities and developmental goals in various age groups, and discuss how these factors often illustrate and explain the observed behavior of children in CRI studies. Similarly, in Section \ref{sec:individual-characteristics}, we analyse how children’s individual characteristics can influence interaction outcomes and consider child-specific tasks that could benefit from a dedicated CRI. In both cases we present prospective experiment designs which support including these age-conditioned and individual difference factors. We postulate that taking these factors into account will not only allow to build a sustainable long-term interaction, but also one that, by being aligned by design with the developmental goals, can effectively engage children and support them in reaching those goals.


\section{Child-Robot Interaction review}
\label{sec:CRI-domains}

Setting the stage for our discussion of Child-Robot Interaction from the combined perspectives of social robotics and child development, in this section we review
the two dominant CRI application domains: healthcare (Sec.~\ref{sec:robots-in-healthcare}) and education (Sec.~\ref{sec:robot-for-learning}). We describe the motivation for using robots in these domains, discuss the problems the robots are typically envisioned to address and outline the main conclusions the research community has reached, in part building on the recent reviews of CRI in healthcare \cite{dawe2019can, papakostas2021social, trost2019socially, van2016robots,moerman2019social,kabacinska2021socially,cabibihan2013robots,ismail2019leveraging,pennisi2016autism} and education \cite{anwar2019systematic,toh2016review,benitti2012exploring,kanero2018social,papadopoulos2020systematic,belpaeme2018social,woo2021use,van2019social}.

\begin{figure}[t]
    \centering
    \includegraphics[width=1.0\linewidth]{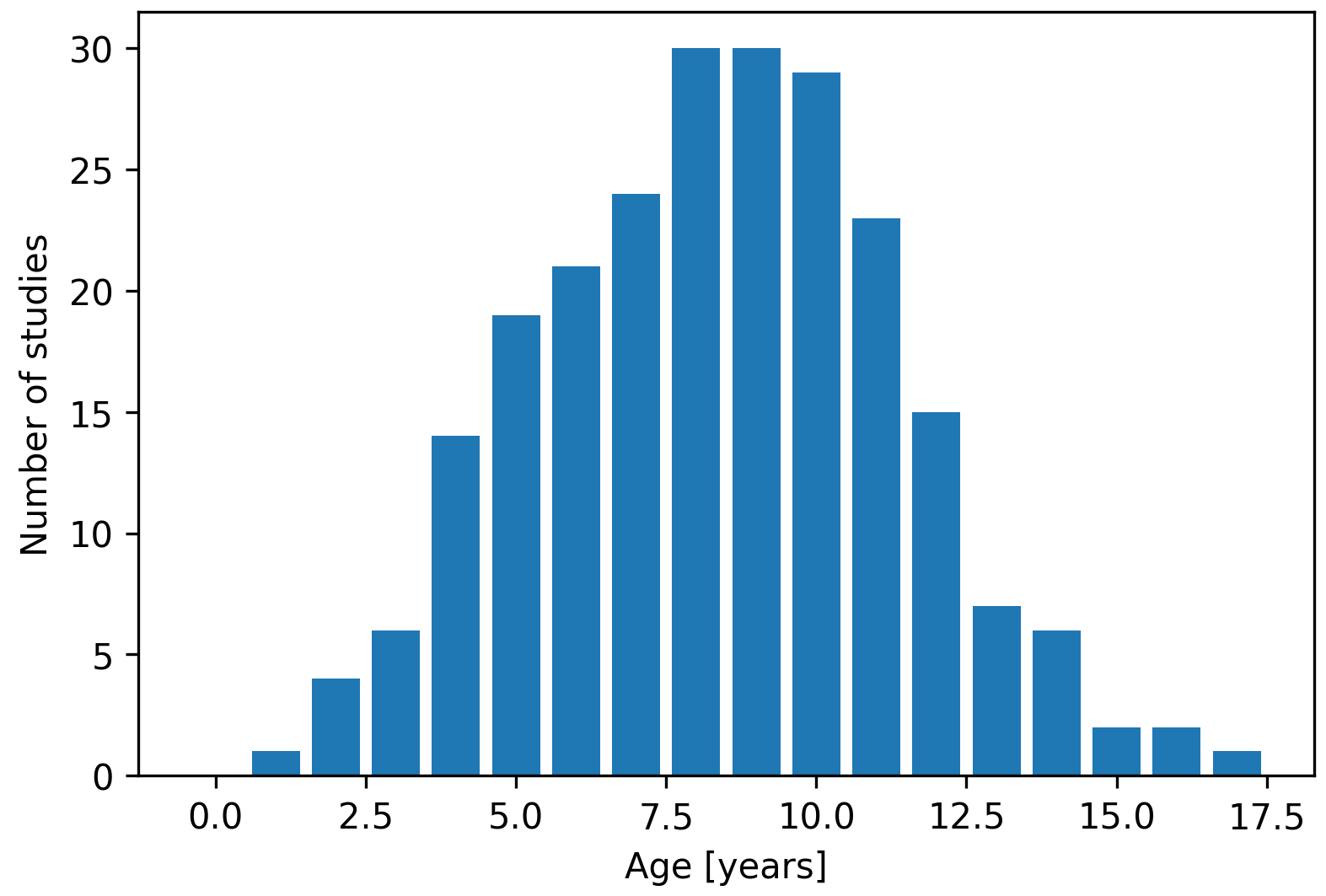}
    \caption{Age distribution histogram, which shows how often children of a certain age participated in the studies from Table~\ref{tab:survey}.
    } \label{fig:age_hist}
\end{figure}

\begin{figure}[t]
    \centering
    \includegraphics[width=0.54\linewidth]{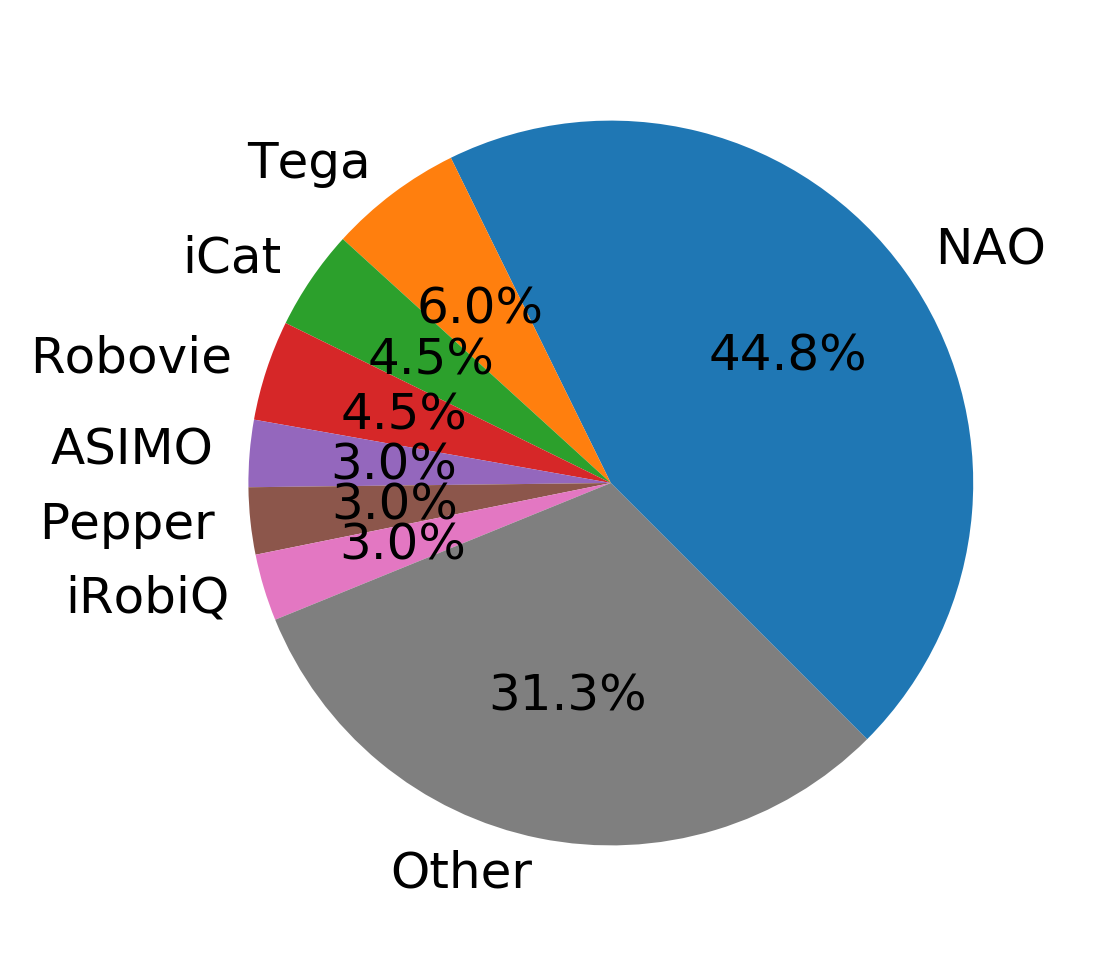}
    \includegraphics[width=0.44\linewidth]{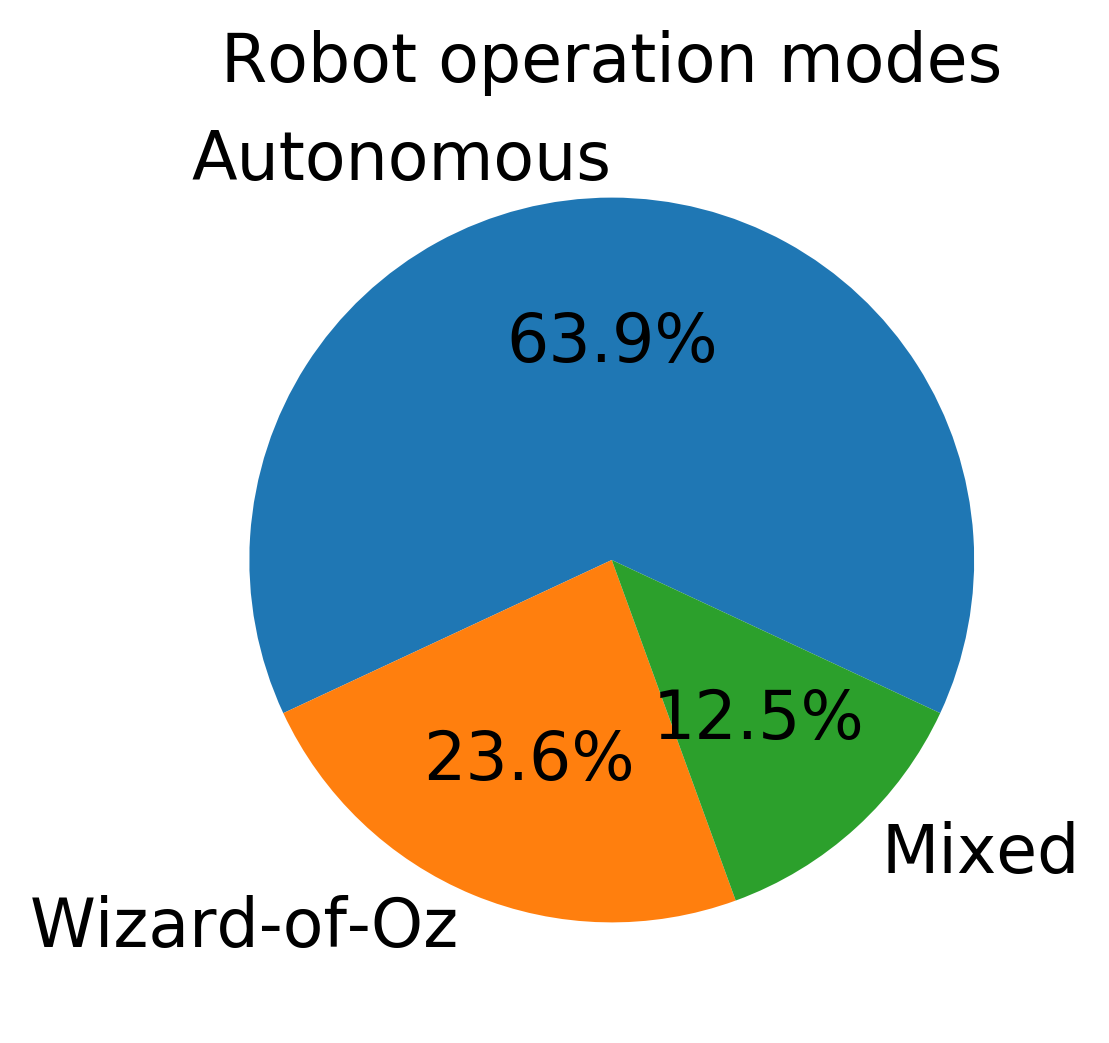}
    \caption{Robots used in the reviewed studies.
    } \label{fig:robots}
\end{figure}

Due to the significant difference to the adult psychology, adult tasks and interactions, in this work we only analyze studies with participants under 18 years of age. Furthermore, we focus mainly on the studies, which provide sufficient information on the participating children and their interaction with the robots to make a psychological interpretation of their behavior and the study outcomes.
In particular, we review the papers which describe:
\begin{itemize}
    \item Children's behavior during their interaction with the robot, e.g. how they engaged, performed in the tasks, participated in games, reacted to the actions and utterances of the robot, etc.
    \item Children's achievements as a result of their interaction with the robot, either reported for groups or single participants, no matter whether they were statistically confirmed or only appeared as tendencies.
    \item Individual characteristics of the children (e.g. collected from interviews, questionnaires, self-reports or parents/teachers), which revealed information relevant to goal of the experiment (e.g. children's interests, preferences, expectations, motivation).
\end{itemize}
A summary of the papers reviewed in this work is presented in Table~\ref{tab:survey}. Relevant statistical conclusions from the review are visually supported by Figures \ref{fig:age_hist}, \ref{fig:interaction_time}, \ref{fig:interaction_modes} and \ref{fig:robots}.

\subsection{Child-robot interaction in healthcare}
\label{sec:robots-in-healthcare}

Healthcare contexts offer a wide range of possibilities for SAR interventions.
Illness can remove children from their normal social networks and pose challenges for coping with treatment, the harmful consequences of diseases and lifestyle changes.
Robots can assist children in managing chronic illness through education and encouragement to perform therapeutic behaviours, help distracting children coping with acute medical procedures, make hospital visits less intimidating or support children’s well-being during long-term hospitalisation and social isolation \cite{dawe2019can, moerman2019social}.

Social robots are also actively employed in rehabilitation as therapy coaches, exercise demonstrators, motivators and progress monitors \cite{dawe2019can}. In this capacity, assistive robots are expected to play a fundamental role for children with all kinds of disabilities, as they extend the ability to play in
children with severe physical disabilities \cite{van2016robots}, facilitate learning for those who suffer from cognitive disorders, engage children with cerebral palsy in exercises to help improve physical functioning \cite{papakostas2021social}.

Some studies in the field of mental health therapy consistently report a number of positive outcomes from the robot intervention, e.g. relief of distress and increase in positive affect in children through distraction and emotional support from SAR \cite{trost2019socially,moerman2019social}. At the same time, the application of robot-enhanced therapy to common mental health problems such as anxiety and depression remains open for investigation \cite{kabacinska2021socially}.

Social robots are traditionally considered to be beneficial in Autism Spectrum Disorder (ASD) interventions, acting as a diagnostic agent, a friendly playmate, a behaviour eliciting agent, or a social mediator \cite{cabibihan2013robots},
aiming at improving the child's social communication skills and reducing repetitive and stereotyped behavior \cite{ismail2019leveraging}.
Robots can stimulate a high degree of motivation and engagement in ASD subjects, including those who are unlikely or unwilling to interact socially with human therapists \cite{scassellati2007social}.
Still, some authors warn against unwarranted enthusiasm \cite{ricks2010trends}. Robot-enhanced therapy provides therapists with a means to more easily connect with the autistic subject \cite{pennisi2016autism}, rather than a treatment to ASD.


Though studies that explore the use of social robots for children in healthcare applications typically report positive outcomes, including generally high acceptance and likability by children, parents, medical staff, teachers and bystanders, many reviewers note that these results should be treated cautiously given the predominance of subjective data \cite{dawe2019can,papakostas2021social,trost2019socially,kabacinska2021socially,pennisi2016autism}.
In particular, many studies focus on case-study, proof of concept, initial response, pilot runs and short-term behavioral assessments \cite{boucenna2014interactive}. The small number of participants, ranged sometimes very broadly by age (e.g. 2--13 or 6--16 y.o.), in combination with the missing control group, are frequent factors that reduce the statistical accuracy of the studies and the reliability of their results. Aiming to investigate acceptability and identify the desired technical characteristics of the robots for different disabilities and cases, many interactions are short-term and single session only \cite{kabacinska2021socially,papakostas2021social} and feature prototype robots, thus limiting replicability \cite{dawe2019can}. 
This is justified as most social robots (and their algorithms) on the market have been developed for children in typical education and may not meet the needs of children with special behavior, perception and emotional reactions \cite{papakostas2021social}.
Overall, the healthcare domain is an uneven and in some parts still emerging field \cite{boucenna2014interactive}, and more studies need to be conducted on a long-term basis (including outside the clinical/experimental context \cite{pennisi2016autism}) in order to prove that robotics can really help children in special conditions or with special needs.


\begin{table*}[p!]
  \centering
  {\scriptsize
  \begin{tabular}{|p{2.985cm}|p{0.575cm}|p{1.66cm}|p{1.065cm}|p{0.39cm}|p{2.82cm}|p{3.11cm}|p{1.27cm}|}
    \hline
    {\bf Paper} & {\bf P. age} & {\bf P. num. / interaction mode} & {\bf Time with R. per S.} & {\bf S. num.} & {\bf Robot / operation mode / role} & {\bf Task} & {\bf Location} \\ \hline
    \rowcolor{Gray}
    Abe et al. 2014 \cite{abe2014toward} & 5-6 & 31 $\diamond$ Japan &  30 min  & 1 & LIPRO*** Peer &  various games with R. & School  \\ \hline
    Ahmad et al. 2019 \cite{ahmad2019robot} & 10-12 & 24 $\diamond$ Australia & 10 -15 min & 4 & NAO** Peer & artificial language ROILA & School \\ \hline
    \rowcolor{Gray}
    Alemi et al. 2014 \cite{alemi2014employing} & 12 & 46 $\diamond\diamond\diamond$ Iran  & 60 min & 10 & NAO* Teach.assist. & L2 vocabulary & School \\ \hline
    Ali et al. 2019 \cite{ali2019can} & 6-10 & 51 $\diamond$ USA & 5 min & 1 & JIBO* Peer & creative activity & After school \\ \hline
    \rowcolor{Gray}
    Alves-O. et al. 2019 \cite{alves2019empathic} &  13-14  & 63 $\diamond\diamond$ Portugal & 30 min & 1 & NAO* Tutor &  collab. game M-Enercities & School \\ \hline
    Alves-O. et al. 2019 \cite{alves2019empathic} &  13-14  & 20 $\diamond\diamond$ Portugal & 30 min & 4 & NAO* Tutor &  collab. game M-Enercities & School  \\ \hline
    \rowcolor{Gray}
    Baxter et al. 2017 \cite{baxter2017robot} & 7-8 & 59 $\diamond$ UK & 5 min & 3-4 & NAO* Peer  & math and stone age & School \\ \hline
    Castellano et al. 2021 \cite{castellano2021pepperecycle} & 7-9 & 51 $\diamond$ Italy & NR & 1 & Pepper* Peer  & recycling education & Lab \\ \hline
    \rowcolor{Gray}
    Chandra et al. 2016 \cite{chandra2016children} & 6-8 & 40 $\diamond\diamond$ Portugal & 15-20 min & 1 & NAO*** Facilitator  & writing improving & School \\ \hline
    Chen et al. 2020 \cite{chen2020teaching} & 5-7 & 59 $\diamond$ USA & 20-30 min & 2 & Tega* Tutor/Tutee & L1 vocabulary & Lab \\ \hline
    \rowcolor{Gray}
    Coninx et al. 2016 \cite{coninx2016towards} & 9-13 & 3 $\diamond$ Italy & 39-57 min & 3 & NAO** Peer & diabetes education & Hospital \\ \hline
    de Wit et al. 2018 \cite{de2018effect} & 4-6 & 61 $\diamond$ Netherl. &  18 min & 1 & NAO* Tutor & L2 vocabulary & School \\ \hline
    \rowcolor{Gray}
    Diaz et al. 2011 \cite{diaz2011building} & 11-12 &  33 $\diamond$ Spain  & NR & 1 & Pleo*/Nao* Peer  & freeplay & School \\ \hline
    Eimler et al. 2010 \cite{eimler2010following} & 9-11 & 18 $\diamond$ Germany & 20 min & 1 & Nabaztag* Tutor & L2 vocabulary & School \\ \hline
    \rowcolor{Gray}
    Elgarf et al. 2021 \cite{elgarf2021once} & 8-9 & 38 $\diamond$ Sweden & 10 min & 1 & Furhat* Peer  & creative activity & Museum \\ \hline
    Fern\'andez-L. et al. 2020 \cite{fernandez2020analysing} &  6-16  & 210 $\diamond\diamond\diamond$ Spain & 60 min & 1 & Baxter* Teacher  & computational thinking & Camp \\ \hline
    \rowcolor{Gray}
    Gordon et al. 2015 \cite{gordon2015can} &  3-8  & 48 $\diamond$ USA & 15 min & 1 & Dragonbot* Tutee  & L1 reading skills & Lab \\ \hline
    Gordon et al. 2016 \cite{gordon2016affective} & 3-5 & 18 $\diamond$ USA & NR &  3-7  & Tega* Peer & L2 vocabulary & School \\ \hline
    \rowcolor{Gray}
    Guneysu et al. 2017 \cite{guneysu2017socially} & 4-12 & 9 $\diamond$ Turkey & NR & 1-2 & NAO* Coach & physical exercises & Lab \\ \hline
    Hashimoto et al. 2011 \cite{hashimoto2011development} & 10-11 & 22 $\diamond\diamond\diamond$ Japan & 30 min & 1 & Saya*** Teacher & science  Leverage principle & School \\ \hline
    \rowcolor{Gray}
    Herberg et al. 2015 \cite{herberg2015robot} & 8-14 & 23 $\diamond$ Singapour & 30 min & 2 & NAO* Tutor & L2 rules & School \\ \hline
    Hindriks et al. 2019 \cite{hindriks2019robot} & 7-9 & 41 $\diamond$ Netherl. & 30-45 min & 2 & NAO* Tutor & math & Lab \\ \hline
    \rowcolor{Gray}
    Hong et al. 2016 \cite{hong2016authoring} & 10-11 & 52 $\diamond\diamond\diamond$ Taiwan & 40 min & NR & Bioloid* Teach.assist. & L2 & School \\ \hline
    Hood et al. 2015 \cite{hood2015children} & 7-8 & 21 $\diamond$/$\diamond\diamond$ Switzerl.  & 8-15 min & 1 & NAO* Tutee & writing improving & School \\ \hline
    \rowcolor{Gray}
    Hsiao et al. 2015 \cite{hsiao2015irobiq} & 2-3 & 57 $\diamond$ Taiwan & 10 min & 8 & iRobiQ* Peer & L1 literacy skills & Kindergarten \\ \hline
    Hyun et al. 2008 \cite{hyun2008comparative} & 4 & 34 $\diamond$ Korea & 25 min & 4 & iRobiQ* Peer & L1 reading skills & Kindergarten \\ \hline
    \rowcolor{Gray}
    Jacq et al. 2016 \cite{jacq2016building} & 5 & 2 $\diamond$ Switzerl. & 40-60 min & 4 & NAO* Tutee & writing improving & Lab \\ \hline
    Jacq et al. 2016 \cite{jacq2016building} & 6-8 & 8 $\diamond$ France & 60 min & 1-3 & NAO* Tutee & writing improving & Lab \\ \hline
    \rowcolor{Gray}
    Janssen et al. 2011 \cite{janssen2011motivating} &  9-10 & 20 $\diamond$ Netherl. &  15 min  & 3 & NAO***  Peer & arithmetic & School \\ \hline
    Jones et al. 2018 \cite{jones2018adaptive} & 10-12 & 24 $\diamond$ UK & 20 min & 4 & NAO* Tutor & geography, map reading & School \\ \hline
    \rowcolor{Gray}
    Jost et al. 2012 \cite{jost2012ethological} & 9-11 & 51 $\diamond$  France & 10 min & 4 & NR** Peer & cognitive stimulation exercises & School \\ \hline
    Kanda et al. 2004 \cite{kanda2004interactive} & 6-7 & 119 $\diamond\diamond\diamond$ Japan & NR & 9 & Robovie* Peer & L2 vocabulary & School/Recess \\ \hline
    \rowcolor{Gray}
    Kanda et al. 2004 \cite{kanda2004interactive} & 11-12 & 109 $\diamond\diamond\diamond$ Japan & NR & 9 & Robovie* Peer & L2 vocabulary & School/Recess \\ \hline
    Kanda et al. 2007 \cite{kanda2007two} & 10-11 & 37 $\diamond\diamond\diamond$ Japan & 30 min & 32 & Robovie*  Peer & freeplay & School/Recess \\ \hline
    \rowcolor{Gray}
    Kennedy et al. 2015 \cite{kennedy2015robot} & 7-8 & 45 $\diamond$ UK & 10-15 min & 1 & NAO* Tutor & prime numbers & School \\ \hline
    Kennedy et al. 2016 \cite{kennedy2016social} & 8-9 & 67 $\diamond$ UK & 11 min & 1 & NAO** Tutor & L2 rules & School \\ \hline
    \rowcolor{Gray}
    Köse et al. 2015 \cite{kose2015effect} & 7-11 & 21 $\diamond$ Turkey & 4 min & 1 & NAO* Peer  & Turkish sign language & Lab \\ \hline
    Konijn et al. 2020 \cite{konijn2020robot} & 8-10 & 86 $\diamond$ Hong Kong & 5 min & 3 & NAO** Tutor & time-tables & School \\ \hline
    \rowcolor{Gray}
    Kory-W. et al. 2017 \cite{kory2017flat} & 4-5 & 45 $\diamond$ USA & NR & 1 & Tega*** Tutor & L2 vocabulary & Lab \\ \hline
    Kose-Bagci et al. 2009 \cite{kose2009effects} & 9-10 & 66 $\diamond$ UK & 6 min & 1 & KASPAR* Peer & drumming & Public event \\ \hline
    \rowcolor{Gray}
    Leite et al. 2014 \cite{leite2014empathic} & 8-9 & 16 $\diamond$ Portugal &  10-25 min  & 5 & iCat* Peer & playing chess & After school \\ \hline
    Leite et al. 2017 \cite{leite2017persistent} &  4-10  & 67 $\diamond$ USA & NR & 1 & PIPER*** Peer  & conversation with R & Lab \\ \hline
    \rowcolor{Gray}
    Ligthart et al. 2022 \cite{ligthart2022memory} & 8-10 & 46 $\diamond$ Netherl. & 15 min & 5 & NAO* Peer  & conversation with R & Lab \\ \hline
    Logan et al. 2019 \cite{logan2019social} & 3-10 & 54 $\diamond$ USA & 30 min & 1 & Huggable*** Peer & freeplay & Hospital \\ \hline
    \rowcolor{Gray}
    Looije et al. 2008 \cite{looije2008children} & 8-9 & 20 $\diamond$ Netherl. & 40 min & 1 & iCat*** Buddy et al. & diabetes education & Lab \\ \hline
    Lücking et al. 2016 \cite{lucking2016preschoolers} & 4-5 & 12 $\diamond$ Germany & 10 min & 1 & NAO* Peer & guessing game & Lab \\ \hline
    \rowcolor{Gray}
    Michaud et al. 2005 \cite{michaud2005autonomous} & 1-2 & 8 $\diamond$ Canada & 4 min & 2 & Roball* Peer & freeplay & Kindergarten \\ \hline
    Nasir et al. 2020 \cite{nasir2020positive} & 9-12 & 78 $\diamond\diamond$ Switzerl.  & 25 min & 1 & QTRobot* Facilitator  & computational thinking & School \\ \hline
    \rowcolor{Gray}
    Neggers et al. 2021 \cite{neggers2021investigating} & 7-11 & 8 $\diamond$ Netherl. & 20 min & 1 & NAO* Peer & diabetes education & Home \\ \hline
    Okita et al. 2011 \cite{okita2011multimodal} & 4-10 & 36 $\diamond$ USA & 20-25 min & 1 & ASIMO*** Peer/lecturer  & table setting game & Lab \\ \hline
    \rowcolor{Gray}
    Okita et al. 2011 \cite{okita2011multimodal} & 6 & 9 $\diamond$ USA & 30-35 min & 2 & ASIMO*/*** Peer & storytelling, knowledge sharing & Lab \\ \hline
    Park et al. 2019 \cite{park2019model} & 4-6 & 67 $\diamond$ USA & NR & 6–8 & Tega* Peer & L2 vocabulary & Lab \\ \hline
    \rowcolor{Gray}
    Ramachandran et al. 2017 \cite{ramachandran2017give} & 8-9 & 38 $\diamond$ USA & 40 min & 1 & NAO* Tutor  & math & School \\ \hline
    Ros et al. 2014 \cite{ros2014adaptive} & 7-12 & 12 $\diamond$ Italy &  8-26 min  &  1-3 & NAO*** Tutor  & dance exercises & Hospital \\ \hline
    \rowcolor{Gray}
    Saerbeck et al. 2010 \cite{saerbeck2010expressive} & 10-11 & 16 $\diamond$ Netherl. & 35 min & 1 & iCat* Tutor  &  artificial language TokiPona & School \\ \hline 
    Sciutti et al. 2014 \cite{sciutti2014you} & 5-18 & 76 $\diamond$ Italy & NR & 1 & iCub* NR & demo and trials at the exhibition & Exhibition \\ \hline
    \rowcolor{Gray}
    Serholt et al. 2016 \cite{serholt2016robots} & 10-13 & 30 $\diamond$ Sweden  & 10-40 min & 3 & NAO* Tutor & map reading & School \\ \hline
    Song et al. 2020 \cite{song2020robot} & 5-12 & 20 $\diamond$ Netherl. & 15 min & 1 & SocibotMini*** Tutor & musical instruments practice & School \\ \hline
    \rowcolor{Gray}
    Tanaka et al. 2007 \cite{tanaka2007socialization} & 1,5-2 & 12 $\diamond\diamond\diamond$ USA & 50 min & 45 & QRIO*** Tutee & free play & Kindergarten \\ \hline
    Tanaka et al. 2015 \cite{tanaka2015pepper} & 4-5 & 10 $\diamond\diamond\diamond$ Japan & NR & 1 & Pepper* Peer  & L2 vocabulary & Lab \\ \hline
    \rowcolor{Gray}
    Tanaka et al. 2015 \cite{tanaka2015pepper} & 3-6 & 17 $\diamond$ Japan & 22 min & 1 & NAO*** Tutee & L2 vocabulary & Lab \\ \hline
    Vinoo et al. 2021 \cite{vinoo2021design} & 2-5 & 6 $\diamond\diamond\diamond$ USA & 60 min & 3 & NR*** Peer & freeplay & Lab \\ \hline
    \rowcolor{Gray}
    Vogt et al. 2019 \cite{vogt2019second} & 5-6 & 194 $\diamond$ Netherl. & 13-21 min & 7 & NAO** Peer  & L2 vocabulary & Lab \\ \hline
    Wright et al. 2022 \cite{wright2022social} & 7-14 & 16 $\diamond$ Switzerl. &  20 min  & 1 & Reachy* Peer & Tower of Hanoi puzzle & Lab \\ \hline
    \rowcolor{Gray}
    Yadollahi et al. 2018 \cite{yadollahi2018deictic} & 6-7 & 22 $\diamond$ Switzerl. & 30 min & 2 & NAO* Tutee & L1 reading skills & School \\ \hline
    Yadollahi et al. 2022 \cite{yadollahi2022motivating} & 8-9 & 22 $\diamond$ NR & 10-15 min & 1 & Cozmo* Guided by child & perspective-taking practice & Lab \\ \hline
    \rowcolor{Gray}
    You et al. 2006 \cite{you2006robot} & 10-11 & 100 $\diamond\diamond\diamond$ Taiwan & 40 min & 4 & Robosapien*** Teach.assist. & L2 & School \\ \hline
    Zaga et al. 2015 \cite{zaga2015effect} & 6-9 & 20 $\diamond\diamond$ Netherl. & 11-18 min & 1 & NAO*** Facilitator & Tangram puzzle & School \\ \hline
    \multicolumn{8}{l}{P. = Participant, R. = Robot, S. = Session, * = Autonomous operation, ** = Mixed operation, *** = Wizard-of-Oz, L1 = First language, L2 = Second language} \\
    \multicolumn{8}{l}{NR = Not reported, $\diamond$ = one-to-one interaction, $\diamond\diamond$ = one-to-pair interaction, $\diamond\diamond\diamond$ = one-to-group/class interaction}
  \end{tabular}
  }
  \vspace{9pt}
  \caption{Summary of the surveyed papers}
  \label{tab:survey}
\end{table*}

\subsection{Child-robot interaction in education}
\label{sec:robot-for-learning}

The first robots to enter the classroom were not meant to be socially assistive, but rather educational tools for teaching engineering and programming. These robots, in addition to positively affecting the learning outcomes, were shown to be highly engaging, motivating and powerful in promoting problem solving skills and teamwork \cite{anwar2019systematic,toh2016review,benitti2012exploring}.

The emergence of Socially Assistive Robots (SAR) extended the application possibilities of robots in education to also support teaching and learning in non-technical subjects. Rather than being a learning tool, robots became participants, collaborators and active members of the learning process \cite{belpaeme2018social}. 

SAR's physical presence and communicative capabilities allow to design various scenarios for the learning process to unfold, enacting different modes of a child’s involvement in it. Robots can be programmed to take up a specific \emph{role}, such as the role of a teacher \cite{hashimoto2011development, fernandez2020analysing} or peer \cite{zaga2015effect}, depending on whether the aim of the learning tasks is to guide students on a task or to have them practice newly learned information with classmates. The role of a tutor is typically applied in one-to-one teaching settings \cite{janssen2011motivating,ramachandran2017give}. Educational benefits can also be obtained with a robot taking the role of a novice, i.e. when the child acts as a teacher for the robot \cite{tanaka2012children,hood2015children,yadollahi2018deictic}.

These roles and their potential are being actively researched in CRI \cite{belpaeme2018social, chen2020teaching,rohlfing2022social}. For instance, some works indicate that in one-to-one interactions the peer role is preferred. In \cite{zaga2015effect} children solved the Tangram puzzle better with a peer robot than with a tutor robot. In \cite{okita2011multimodal} children progressed in a table setting task better when paired with a ``cooperative interaction style'' robot, as compared to a ``lecture interaction style'' robot.
The learning buddy role (as compared to motivator and educator) was shown to be the most supportive in healthcare related activities, such as adherence to diet and exercises for diabetics \cite{looije2008children}.
Conversely, when interacting with a group of learners, the robot seems to achieve more positive outcomes in the role of a teacher or teacher's assistant \cite{you2006robot, hong2016authoring, alemi2014employing} or facilitator \cite{chandra2016children, nasir2020positive}. Lastly, the role of a novice is considered best to support and engage children with weaker skills, as e.g. it was reported for improving handwriting \cite{hood2015children,jacq2016building} or reading skills \cite{yadollahi2018deictic}.

In teaching settings in one-to-one interactions, typically acting as tutors, social robots help realising the personalized strategy of teaching, which implies flexibly tuning the robot's social behavior to match the ongoing interaction in order to support the task engagement, adequately react to the affective state of the child, or adapt a teaching strategy to match the learning progress.
Some studies show that a personalized (adaptive) tutor can increase engagement and promote learning by properly timing the breaks \cite{ramachandran2017give}, giving adaptive emotional feedback \cite{ahmad2019robot}, determining a personal learning goal instead of a pre-defined group target \cite{janssen2011motivating}, employing adaptive training, optimized for a child’s engagement and skill progression \cite{de2018effect,park2019model}, modifying tasks to align them with the child performance \cite{baxter2017robot}, teaching by a social dialog \cite{saerbeck2010expressive}, or adaptively switching between the tutor/tutee roles \cite{chen2020teaching}. 
At the same time, however, other studies did not report significant effects of the personalized 
interactions \cite{hindriks2019robot,jones2018adaptive,gordon2016affective,kory2017flat, kennedy2016social, alves2019empathic,baxter2017robot}.

From a content perspective, the main body of evidence on deploying robots for educational purposes comes from the realm of first and second language learning \cite{kanero2018social,papadopoulos2020systematic}, see also Table~\ref{tab:survey}. Further applications for SAR are also explored in the fields of tutoring musical \cite{song2020robot} and physical skills \cite{guneysu2017socially,litoiu2015robotic}, dance \cite{ros2014adaptive}, handwriting improvement \cite{chandra2016children,hood2015children,jacq2016building}, spatial skills practicing \cite{yadollahi2022motivating}, teaching math \cite{baxter2017robot,hindriks2019robot,janssen2011motivating,kennedy2015robot,konijn2020robot}, computational thinking \cite{fernandez2020analysing, nasir2020positive}, history \cite{baxter2017robot} and geography \cite{jones2018adaptive}.

When evaluating the efficacy of SAR intervention in education, the most commonly reported result is a positive affective outcome \cite{papadopoulos2020systematic}. 
Robots make the education process significantly more engaging and attractive, with many children, upon their participation in studies, signalling their interest to continue learning with the robot.
Since the goal of using robots in education is primarily increasing the learners' academic performance, and given that engaged students would interact with the given task for longer, it is natural to expect a rapid learning progress as an outcome of educational CRI. However, the relation between the affective and cognitive outcomes is not as straightforward. Positive affective outcomes from learning with the robot do not necessarily imply positive cognitive outcomes, and vice versa \cite{belpaeme2018social}.
For instance, in some studies, children being taught with robot assistance (contrary to studying without a robot) increased  achievements in learning new words \cite{alemi2014employing, kennedy2016social, tanaka2007socialization}, listening and reading skills \cite{hong2016authoring}, drumming performance \cite{kose2009effects}. Conversely, in other studies, children learned vocabulary equally well when paired with a robot or with another student, but not better than those children who learned individually using a computer program or a dictionary \cite{eimler2010following,vogt2019second}. 
In \cite{nasir2020positive}, students achieving low and high performance in the learning tasks reported equally positive perception of the robot (including its intelligence, likeability etc.) and a similar positive assessment of its helpfulness.
Lastly, some authors report that the robot might counter-productively affect the learning process, distracting the student from the task \cite{kennedy2015robot, herberg2015robot,yadollahi2018deictic, konijn2020robot}.

\begin{figure}[t]
    \centering
    \includegraphics[width=0.85\linewidth]{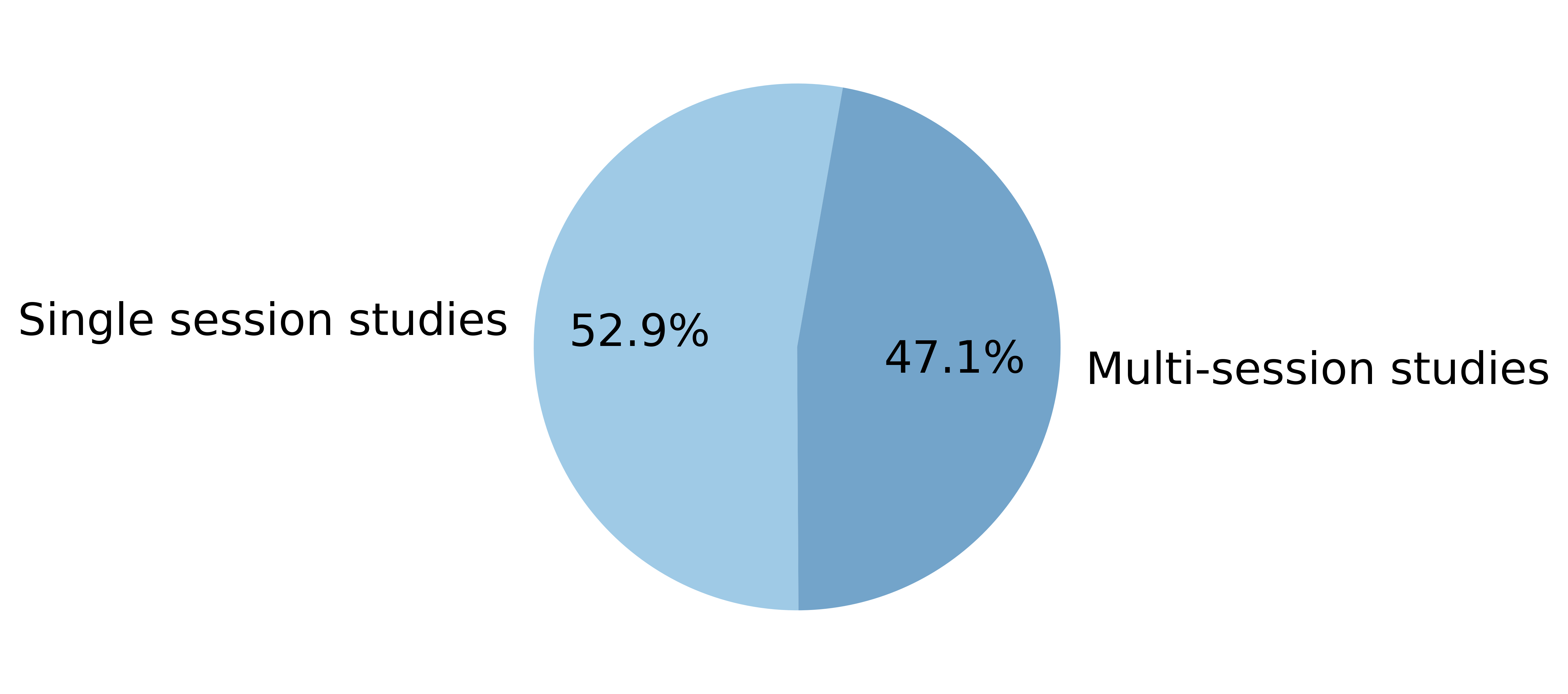} \\
    \includegraphics[width=0.51\linewidth]{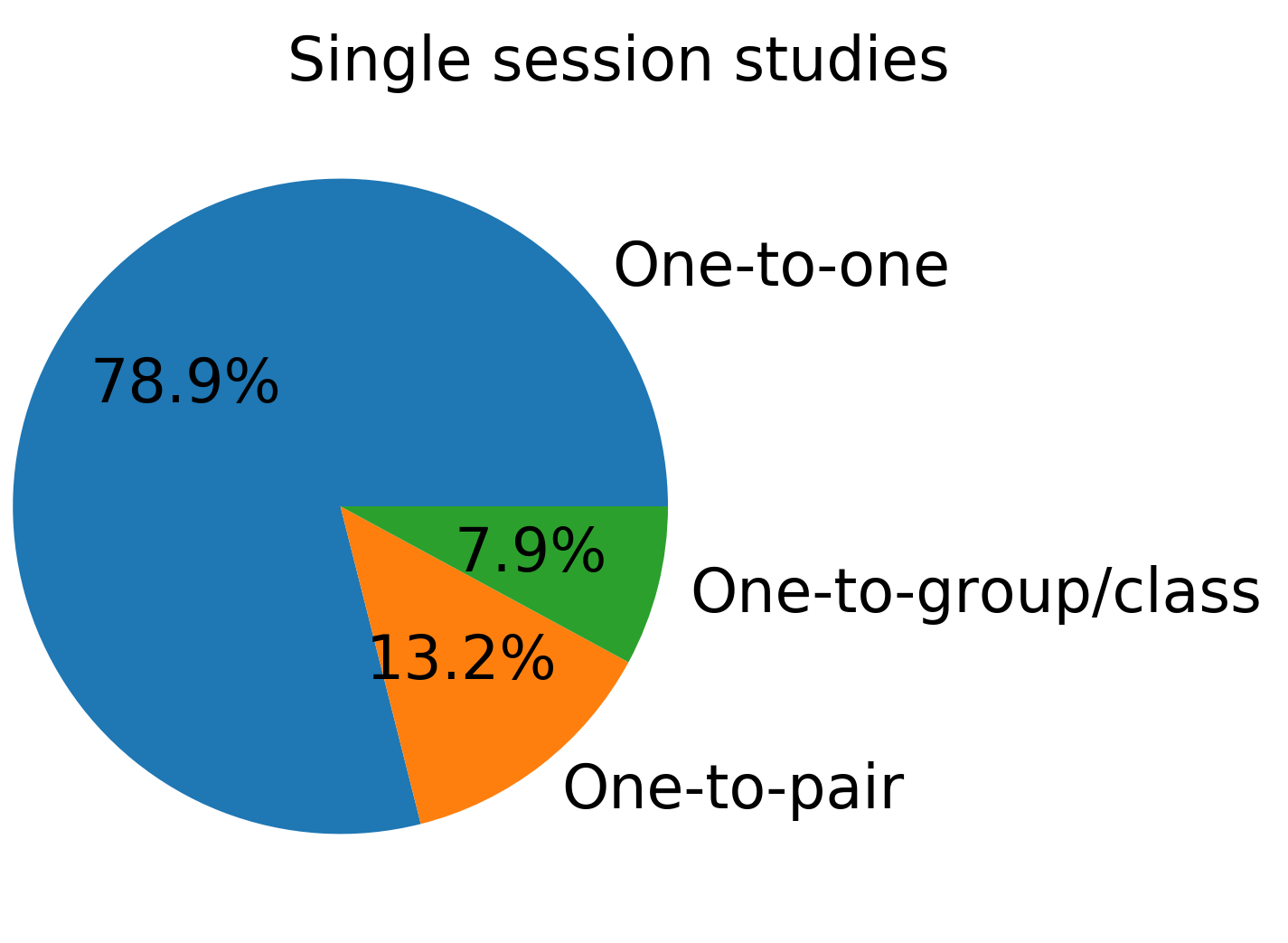}
    \includegraphics[width=0.48\linewidth]{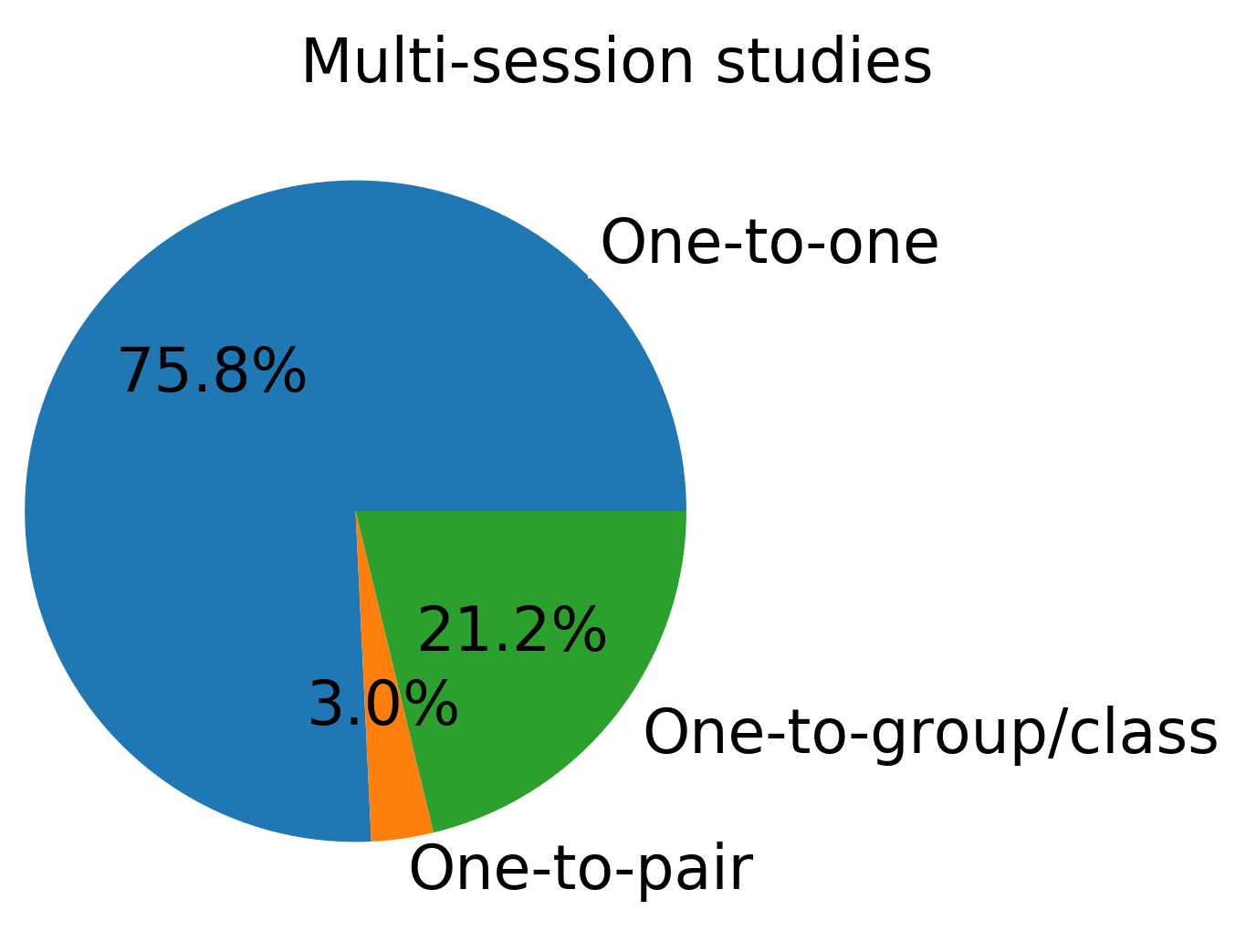}
    \caption{Interaction modes in the reviewed studies, shown separately for the single-session and multi-session scenarios.
    } \label{fig:interaction_modes}
\end{figure}

In general, the effect of a robot's interventions in learning tasks is not sufficiently and conclusively described by the existing studies. According to the recent review by Johal \cite{johal2020research}, most of the works reporting cognitive outcomes present results on immediate post-tests, while only 15\% also report on retention outcomes. The majority of studies are conducted as individual tutoring, leaving a gaping question of how to sustain the learning process in diverse educational situations (see Fig.~\ref{fig:interaction_modes}).
There are many signs that the robot might facilitate the learning progress of the child, but researchers still face the challenge
to design, develop and fine-tune interactive robots that would consistently yield learning gains, which is precisely what should be the most important evaluation metric.

\begin{figure}[t]
    \centering
    \includegraphics[width=1.0\linewidth]{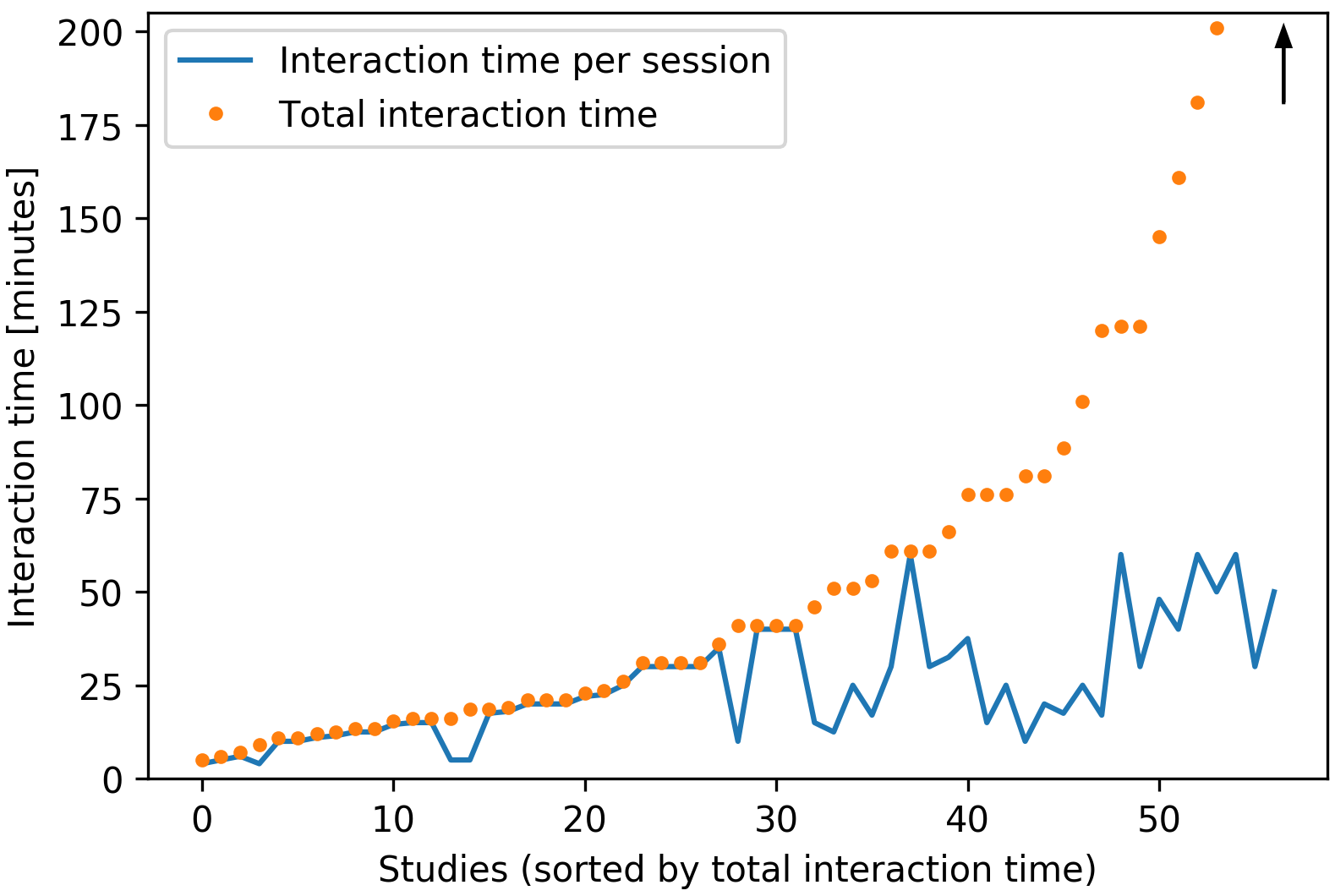}
    \caption{Interaction time with the robot in one session (blue line) and cumulatively over multiple sessions (``Total interaction time'', orange dots). This figure illustrates the studies from Table~\ref{tab:survey}, ordered by how long children interacted with the robot. The last three values in the total interaction time plot (600, 960 and 2250 minutes, indicated by the black arrow) are omitted for better readability. As the graph shows, nearly half of the studies envision a single session, lasting max. 25 minutes, while longer ones envision multiple sessions, each lasting up to 50 minutes.
    } \label{fig:interaction_time}
\end{figure}


\section{Towards sustainable long-term interactions}
\label{sec:robot-for-a-long-term-interaction}

The tasks of an assistive robot, designed to be embedded in a child's daily activities, likely will not be restricted to isolated interactions. This is equally true for all domains, e.g. learning, rehabilitation, physical and cognitive development, social activity etc.
The need for social support is usually not resolved within one interaction session, and, similarly, other complex goals of the child's development and well-being require long-term interventions \cite{kabacinska2021socially,gargot2021not}.
The CRI design therefore should enable, facilitate and encourage long-term interactions.
Many authors raise this point \cite{baxter2016characterising,coninx2016towards,kabacinska2021socially,woo2021use}, but in fact most of the studies focus on short-term and single-session interactions, as we noted in Sec.~\ref{sec:CRI-domains}, see also Table~\ref{tab:survey} and Fig.~\ref{fig:interaction_time}. Beside logistics and budgetary constraints, this limitation arises from the technical and AI capabilities of the robots, as well as the complexity of maintaining a child's interest and attention towards working with the robot in the long term.

The technical limitations of state-of-the-art SAR only enable the robot to perform restricted tasks in restricted contexts \cite{belpaeme2018social}, which naturally leads towards the development of short-term interaction scenarios, primarily tailored to and often investigating what the robot can do, as we noted in Sec.~\ref{sec:robots-in-healthcare}.
As a consequence, the repertoire of activities the robot can offer to children is limited and predictable, unable to support engagement over extended periods of time.
Using semi-autonomous robots, for instance in Wizard-of-Oz settings, may extend the range of activities, but ultimately shifts the focus of the studies from CRI to interactions between the adults behind the robot and children, with the robot merely acting as a proxy.
Indeed, there is a solid consensus in the community around the opinion that a socially-assistive robot interacting with children should be fully autonomous in order to reach its long-term goals \cite{salter2008going,lucking2016preschoolers,ros2011child, baxter2016characterising, ahmad2019robot, coninx2016towards,wright2022social,dawe2019can}.
While in recent years more and more autonomous socially assistive robots are being developed and brought to test \cite{johal2020research},
maintaining the long-term interest of children is still a major challenge.

A key reason for this fact is that the majority of children are still not used to interacting with robots and therefore their experience and perception of the robot are greatly affected by both its novelty and fun effect.
As outlined in Section \ref{sec:CRI-domains}, children tend to enthusiastically engage in any activity involving a robot, whether it is learning new words or solving a puzzle, practicing dancing or perspective-taking, mobility improvement or cognitive skills exercises.
Their enthusiasm towards experiencing the new technology is so high, that they try to solve the task the best they can, reaching results which wouldn't have been attained had the learners been more familiar with the robot \cite{liu2009impact}.
As the novelty effect wears off, the results tend to stop growing and the engagement with the robot declines almost to becoming negligible \cite{leite2013social, you2006robot, serholt2016robots,papakostas2021social}.

Indeed, the novelty of the robot is a double-edged sword for CRI.
On the one hand, it is a powerful attractive factor for most children, who expect an extraordinary interaction with the robot.
As discussed in Section \ref{sec:CRI-domains}, this factor is fundamental for many successful applications, especially in the healthcare domain, aimed at emotional support or distracting from distressing experiences \cite{trost2019socially,moerman2019social,kabacinska2021socially}. The novelty of the robot also serves as an initial motivator for the children to get involved in learning tasks, even ones they would otherwise not find very attractive\cite{papadopoulos2020systematic}.
The other edge of the metaphorical sword is that the novelty effect typically wears of quickly and the robot needs other ways to engage children into long-term interactions. In addition, the novelty effect complicates the research and development of effective CRI models, as it is difficult to separate its impact on the outcomes from the robot's own contribution. Short and early sessions are thus not reliable predictors of how the interaction would progress in the long-term.

Personalization, as we discussed in Sec.~\ref{sec:robot-for-learning}, is a promising and popular research direction for sustaining engagement beyond the short term. While this seems to be particularly important in educational contexts, other activities have also shown to be prolonged thanks to personalization: an empathic robot acting as a companion in chess can extend the play for a longer period \cite{leite2014empathic}, and a robot that memorises previous dialogues can keep children engaged for longer in conversations \cite{leite2017persistent, ligthart2022memory}.


And still, personalization alone is likely not sufficient to achieve long-term engagement, even in fully-autonomous and technically-advanced robots. The key insight is that personalization attempts to catch and captivate an ongoing interaction, i.e. working around the consequences and not looking into the causes for the interaction to unfold as it does. We argue that these causes could be revealed when reviewing the interaction scenario from the child psychology and development standpoint, i.e. considering the child factor in CRI.

To this end, in the next section we review the state-of-the-art CRI scenarios introduced in Section~\ref{sec:CRI-domains} from the child-centered perspective, highlighting the key role that the child factor plays for the success of the interaction. We show that setting child-appropriate goals (e.g. practicing a skill, investigating the objects and their properties, achieving academic targets) and format (e.g. conversation, lecture, game, creative task) of the interaction is as important for its success as the captivating design of the robot and its personalization capabilities.
We postulate that considering these factors can yield fundamentally more engaging and sustainable interactions, thus increasing the long-term efficacy of CRI and the potential to reach the interaction goals.

\section{The child factor in CRI}
\label{sec:robot-for-the-child}

The analysis of an interaction begins with defining its subject, i.e. the goal, towards which the efforts of the involved parties are directed. The subject of CRI is naturally the task or activity that the child is addressing together with the robot. This task can be defined by parents, teachers, therapists, or the child him- or herself. The result of the interaction will, accordingly, depend on the efforts and mood of the child -- the satisfaction from getting better at something, the disappointment at not succeeding, the interest, engagement and enthusiasm towards the activity, or the urge to switch it or give up on the robot.

Even though the child's progress and well-being is the central point of CRI, the interaction scenario is often selected as a result of robot-centric considerations, or at least greatly affected by them. As a consequence, its effectiveness and sustainability depend primarily on the child making something out of it, which in turn depends on his or her individuality, capabilities, interests and background.
These factors are usually considered post factum (e.g. to evaluate the experience with the robot in questionnaires), but rarely used to design or drive an interaction: indeed, the robot's behavior is typically adapted using only the recognized emotions and responses, as we discussed in Sec.~\ref{sec:robot-for-learning}.
This approach reduces the model of a child to a generic black box,
and no assumptions are made on how the interaction will proceed from his or her perspective. The offered tasks or activities might turn out to be exciting, boring or unclear, and the interaction may stall after the novelty effect wears off for reasons related to the task itself rather than the robot or its behavior.
In this section we review several studies
where the observed interactions and the task progress to a large extent depended on the individual differences between the participants.

\begin{figure}[t]
    \centering
    \includegraphics[width=0.575\linewidth]{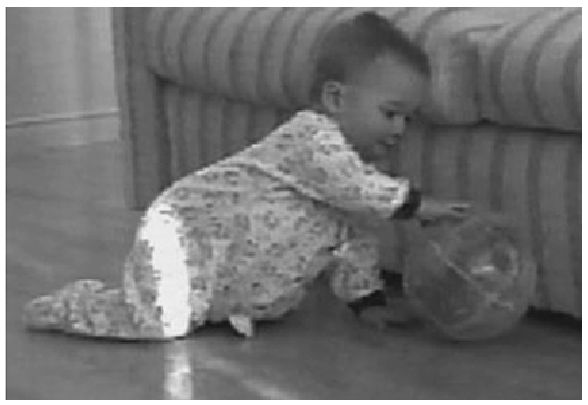}
    \includegraphics[width=0.39\linewidth]{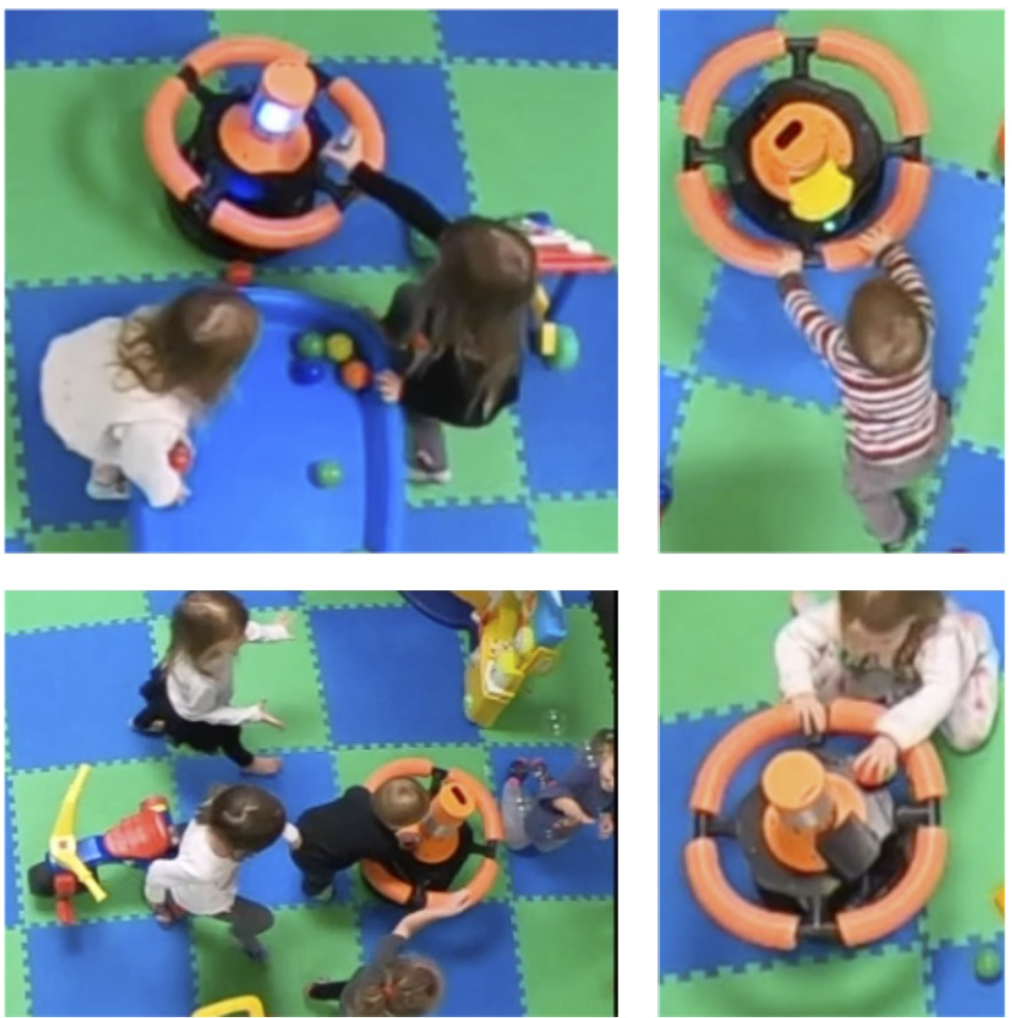} \\
    \includegraphics[width=0.53\linewidth]{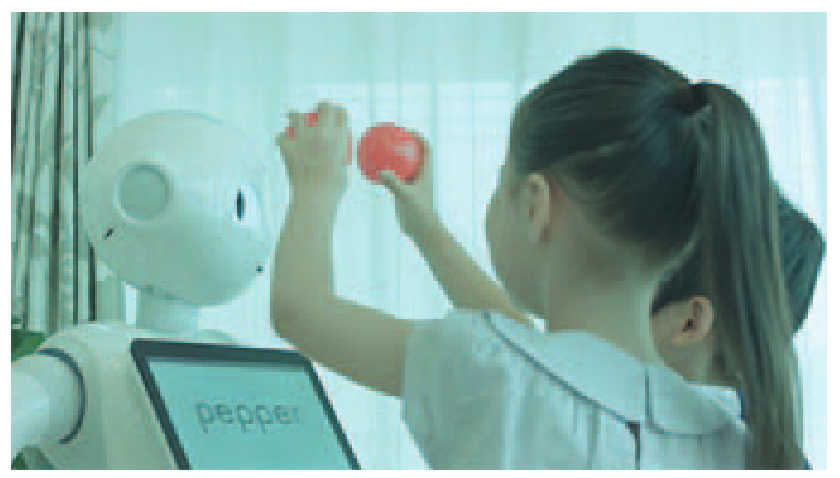}
    \includegraphics[width=0.44\linewidth]{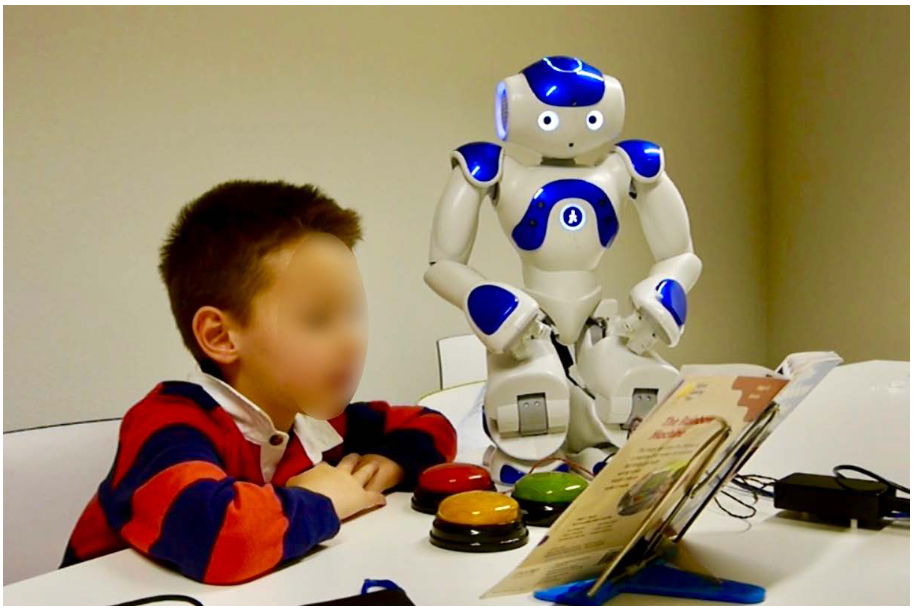} \\
    \includegraphics[width=0.97\linewidth]{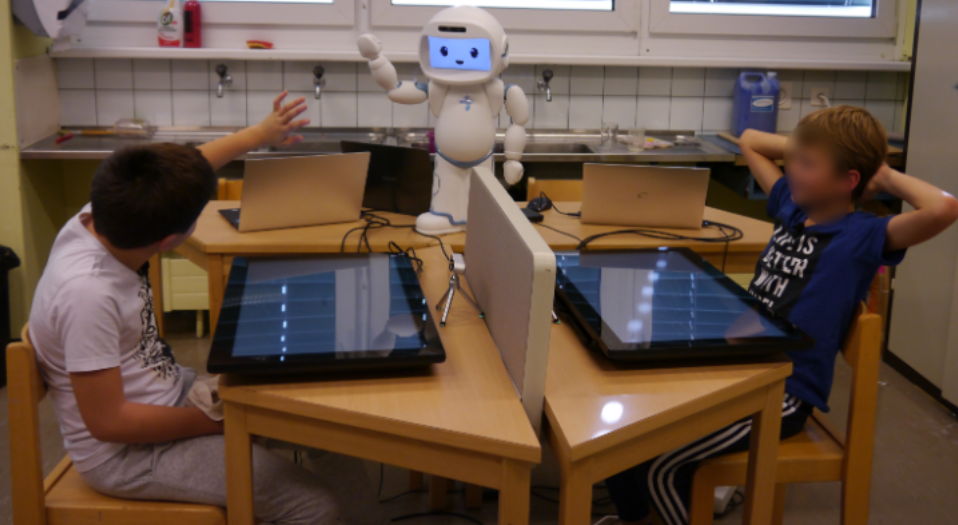}
    \caption{Illustrations from some of the child-robot interaction studies reviewed in this work. {\bf Top left}: Michaud et al. 2005 \cite{michaud2005autonomous} (``Roball''), {\bf top right}: Vinoo et al. 2021 \cite{vinoo2021design}, {\bf middle left}: Tanaka et al. 2015 \cite{tanaka2015pepper}, {\bf middle right}: Yadollahi et al. 2018 \cite{yadollahi2018deictic}, {\bf bottom}: Nasir et al. 2020 \cite{nasir2020positive}.
    } \label{fig:cover}
\end{figure}

Kanda et al. \cite{kanda2004interactive,kanda2007two} show how individual differences between children may affect the duration, the format and the results of the interaction.
In \cite{kanda2004interactive} two groups of children (6-7 and 11-12 y.o. respectively) had the opportunity to freely play with an English-speaking robot, thus improving their ability to speak English. After the first week most of the children lost their interest in the robot, and there was no overall learning progress among them. A few children, however, continued the communication with the robot and improved their English level by the end of the second week.
Comparing the background of these few children with the rest of the participants, authors revealed that having some initial interest in English or already knowing a little bit of English helped to sustain the long-term interest to the robot. 

The authors further highlight the effect of individual differences in the follow-up work \cite{kanda2007two}.
Here the robot attempted to captivate the interest of all the children in the group (10-11 years old) by addressing them in a personalized manner, for instance by learning their names and recognizing them among the classmates. The robot also introduced new interactive behaviors (such as shaking hands, hugging, playing rock-paper-scissors, exercising, singing, and pointing to an object in the surroundings), once the total duration of each individual interaction surpassed a certain threshold (e.g. 120 minutes). Additionally, the robot increasingly often made personal remarks, with comments such as ``I like chattering'' (the robot tells this to a child who has played with it for more than 120 minutes), ``I don’t like the cold'' (180 minutes), etc.
The robot successfully continued interacting with many children for two months (32 sessions of free play, each 30 minutes long).
As in \cite{kanda2004interactive}, the participants' engagement decreased after an initial excitement, but increased again in the last two weeks, when the children were told that the robot would soon go away.
Interestingly, a small group of children interacted with the robot longer than others (10 out of 37 participants). The children characterized themselves on three scales: ``want to be friends with the robot'', ``want to know the mechanism of the robot'', ``usually play indoors''. These three characteristics turned out to define the level of involvement: the more a child wanted to be friends with the robot and the less the child wanted to know its mechanism, the longer he or she played with the robot. Additionally, the children who usually played indoors, interacted with it longer, simply because the robot was also indoors.

By calling their approach ``pseudo-development'' \cite{kanda2007two}, Kanda et al. highlight a key aspect of sustainable CRI. As long as the interaction scenarios are pre-defined, their natural ``development'' is limited to the scripted content. The robot may offer the child to engage in certain activities, but eventually the success of the interaction will depend on the child.
In \cite{kanda2004interactive} and \cite{kanda2007two}, the robot offered every child to come by and communicate, and all children engaged in this activity to some extent over the course of two months.
But every child had an individual interest in the robot, which eventually defined the degree of his or her involvement. Apart from being ``interested in the mechanism'', and ``wanting to be friends'', some children played with the robot simply because the robot was ``indoors''. Curiously, in \cite{kanda2004interactive} two children continued interacting with the robot out of compassion, because no one else did.

This indicates that children invent their own interaction scenarios, as also observed by Michaud et al. \cite{michaud2005autonomous}. In this study, 8 children aged 12 to 24 months interacted with a Roball (self-propelled ball-shaped socially-assistive robot, see Fig.~\ref{fig:cover} top left).  
Each of the 8 children, as the authors note, adopted a different interaction style with Roball: some were interested in moving Roball, some were attracted by Roball's face, some imitated the Roball's rolling actions, some attempted to stop it, etc.
Even at such a young age, children have their own personalities and interests and may not be at the same development stage even if they were the same age.

It is worth stressing here the importance of a child's own internal reasons towards defining their interaction with a robot, since many studies are instead predicated on the hypothesis that the effectiveness and sustainability of long-term CRI strongly depends on the robot’s ability to elicit engagement in children.
Under this hypothesis, it is assumed that a robot with sufficient social and cognitive skills would successfully persuade the child of its competence as a companion, teacher or therapist, and make the child feel important for the robot \cite{diaz2011building,leite2013social}. In short, it is assumed to be possible to elevate the child-robot interaction (in terms of quality and desired outcomes) by improving the behavior of the robot \cite{coninx2016towards}.
Contrary to this assumption, literature points in the direction that no amount of robot competence or skills will likely be sufficient to compensate for unique factors of the individual child. Consequently, as the above examples illustrate, different types of interactions, varying levels of engagement, contrasting outcomes will most likely always emerge from interactions between a robot and different children, if individual factors are not properly considered.

Taking individual factors into account is important not only to increase the number of children who would engage in the interaction, but also the effectiveness of the interaction. 
Indeed, while children are typically always satisfied with the robot, as they enthusiastically report in the post-experimental questionnaires, the interaction is often not effective in achieving its desired outcome.
Figuring out the robot's capabilities turns out to be a challenging task.

Konjin and Hoorn \cite{konijn2020robot} show that even a small insight on the child's background may help building an efficient interaction.
In their study children 8 to 10 years old rehearsed the time tables together with the robot. The study investigated how the behavior of the robot (social or neutral) affects the learning outcomes of children with different academic abilities (above-average and below-average).
The results show that the above-average pupils profited equally from both conditions of the robot, whereas below-average pupils benefited more from the robot that showed neutral behavior.
The authors conclude that the social robot might distract the children, especially the less academically apt. The neutral robot may therefore help these children concentrate better and thus increase the chances of a successful interaction.

The fact that collaborative learning with a robot may not be beneficial due to children’s individual differences also finds evidence in the study by Yadollahi et al. \cite{yadollahi2018deictic}, in which the robot was supposed to improve children's reading performance by bringing their attention to the text with deictic gestures. Surprisingly, the robot's gesture support was more beneficial for children who already had high reading skills. Students who were not as proficient in reading benefited from the robot's pointing in easy-level tasks, but performed better without robot's pointing in difficult tasks. The authors note that, in difficult tasks, the robot's pointing distracted the children from the text and proved to be counterproductive. Therefore, how the robot's pointing could contribute to improving the performance of students with low reading skills remains an open research question.


Contradictory results on the effect of personalized robot behavior were also reached in the study by Baxter et al. \cite{baxter2017robot}. Here, personalized behavior did not have an effect in mathematical assignments -- a subject the participants were previously trained in, while they seemed to benefit the children in a new subject (``stone age history'').
This disparity could be attributed to the individually different prior training and personal interest into the subjects (mathematics and stone age history). Without background information on the study participants, it is not possible to properly identify the cause of the observed effects.

Contradictory outcomes are also observed in CRI studies across various age groups. Many authors note that children of diverse ages respond differently to the same interaction with a robot. For instance, younger children are more likely to engage with robots, while older children expect more complex, diverse and dynamic behavior from the robot \cite{leite2013social}. Younger children are more lenient to the robot's imperfections and errors \cite{hood2015children}, whereas older children appear more apprehensive or cautious around robots and technology \cite{fernandez2018may}.
An analysis of how children perceive the robot’s properties suggests that kids until 9 years of age give more relevance to a humanlike appearance, while older children pay more attention to the robot action skills. Furthermore, researchers note that the possibility to see and interact with a robot has an impact on children’s judgments, especially convincing the youngest to consider also perceptual and motor abilities in a robot, rather than just its appearance \cite{sciutti2014you}.

While these observations confirm that age is an important factor for CRI studies, the question of \emph{why} children in certain age groups interact with the robot differently from others needs to be explored further.
Age groups themselves are often vaguely justified and broadly defined, for instance
to include children from 3 to 8 years old \cite{gordon2015can} or 5 to 16 years old \cite{beran2011understanding}, which clearly means very different individuals on various stages of development.

In conclusion, while some authors notice and explore the child factor in the CRI studies, in most cases the child remains a rather abstract exogenous factor, which impedes explaining and predicting the observed interactions and the effects of the socially-assistive robots.
In this section we highlighted that children may have their own reasons to interact with the robot, invent their own interaction scenarios, as well as interpret the robot’s intentions in their own way and respond to them contrary to the researchers' expectations (e.g. being distracted instead of concentrating on the task).
We argue that this behavior may occur for various reasons that have nothing to do with the operation of the robot.
In the following Sec.~\ref{sec:age} and \ref{sec:individual-characteristics} we proceed to describe the factors along which differences between children emerge, namely development level predicated on age and individual characteristics.

\section{Age differences}
\label{sec:age}

\subsection{Development stages}
\label{sec:development-stages}

Child development does not fit rigid age thresholds, but rather advances incrementally, continuously and individually. Still, it is possible to define certain ``stages'', conditioned on the age but primarily defined by the biological maturity of the organism, uneven development of the psyche and the accumulated social and cognitive experiences of a particular individual. These stages are separated by developmental crises (at the age of one, three, six-seven years old and puberty), which sometimes resolve unobserved and in a calm manner, and sometimes may cause a stark conflict with parents and the environment.
The development of younger children can be accurately explained and predicted with general age-based trends, whereas the importance of past experiences and individual characteristics becomes more critical with age, causing noticeable differences among similarly aged older children \cite{craig1999human}.
Furthermore, in the field of developmental science, there exist diverse perspectives regarding the stages' continuity and discontinuity in child development. For a more in-depth exploration of this subject, we recommend the book by McCormick and Scherer \cite{mccormick2018child}.

Moving forward, our objective is to outline more precisely the expectations associated with children of specific age groups and to elucidate the implications of this understanding for the design of interactions with children. In crafting a comprehensive understanding of the various stages, we will draw upon the insights provided by prominent developmental theories: Erikson’s Theory of Psychosocial Development  \cite{erikson1993childhood}, Piaget's Theory of Cognitive Development \cite{piaget2003psychology}, Vygotsky's Theory of the Higher Mental Functions \cite{vygotskiui1997collected}, Elkonin's Theory of Dominant (Leading) Activity \cite{el2017toward}.
These theories form the foundation of knowledge in child development, mastering which is the basis of education in the field \cite{siegler2020how,mccormick2018child, shaffer2013developmental}. 

\begin{table*}[t]
  \centering
  \begin{tabular}{|p{2.9cm}|p{2.9cm}|p{3.7cm}|p{2.6cm}|p{3cm}|}
    \hline
    & {\bf Stage of psychosocial development (Erikson)} & {\bf Stage of cognitive development (Piaget)} & {\bf Dominant mental function (Vygotsky)} & {\bf Dominant (leading) activity (Elkonin)} \\
    \hline
    Infant (0-1 y.o.) & Trust & Sensorimotor stage (up to 2 y.o.) & Attention & Direct emotional contact with the adult \\
    \hline
    Toddler (1-3 y.o.) & Autonomy & Sensorimotor stage (up to 2 y.o.) Preoperational stage (2-7 y.o.) & Perception & Manipulation of objects \\
    \hline
    Preschooler (3-6 y.o.) & Initiative & Preoperational stage (2-7) & Memory & Role playing \\
    \hline
    Gradschooler (6-12 y.o.) & Industry & Concrete operational stage (from 7 to 11-12) & Concrete logical thinking & Formal learning (schooling) \\
    \hline
    Adolescent (12-18 y.o.) & Identity & Formal operational stage (from 11-12 and above) & Abstract logical thinking & Intimate personal relations within the peer group \\
    \hline
  \end{tabular}
  \vspace{9pt}
  \caption{Developmental stages of a child and their main properties}
  \label{tab:developmental-stages}
\end{table*}

In Table~\ref{tab:developmental-stages} we summarize the child developmental stages. 
Below follows a short description for each stage (for more detailed information we also recommend \cite{stern2017psychology, nakkula2020understanding,louw2014child}).

\begin{itemize}

\item {\bf Infant} (Up to 1 year):

The key task in infancy is the development of trust towards the surrounding world, which is formed during a direct emotional contact with the adult. This short period encompasses a dramatic transformation of activity, beginning with the inherent basic reflexes (sucking, grabbing, watching, crying) and leading to intentional actions and the ability to deliberately communicate with people around. Attention as a dominant function develops from fixating and following at random to a voluntary process. Cognitive functions develop on the base of sensorimotor and self-motion experience. At 8-9 months the infant is able to crawl, reaching an important milestone in independence. 

\item {\bf Toddler} (1-3 years old):

Once able to walk, the toddler begins actively exploring the areas around. This is a stage of increasing autonomy and independence. The dominant mental function at this stage is perception, leading to the active accumulation of sensory experience. The toddler studies the world and the properties of objects by directly manipulating and sensing them, and develops memory to recognize these properties in other objects (e.g. small and large, light and heavy, quiet and loud). Mimicking the adults, basic speech forms, and other sorts of imitation appear in behavior and actions with toys.

\item {\bf Preschooler} (3-6 years old):

This is the stage of initiative. The child learns to distinguish him- or herself from others, actively indicating own wishes and borders, which is supported by rapid speech development, as well as the accumulation of vocabulary. Memory as a dominant mental function contributes to memorising a large amount of information, but the child struggles to logically process it. This leads to limited, often ``magical'' explanations of causal relations, grounded in the specific (not generalized) instances of experience. The cognitive function is largely egocentric, animistic and anthropomorphic. The leading form of activity at this age is role playing with peers and imaginary characters. By acting various social roles, children explore and learn to behave according to the social conventions of the society.

\item {\bf Gradschooler} (6-12 years old):

This period is marked by rapid intellect development and industry, largely driven by the school activities and formal learning. The importance of achievements in various subjects in and outside of school increases, and the sense of competence forms. Logical thinking becomes dominant as the child learns to model and explain causal relations which go beyond immediate sensing. Nevertheless, gradschoolers still struggle with abstract and hypothetical concepts, and require an effort to apply the logical reasoning to concrete experiences.
Behavior becomes increasingly deliberate and self-controlled by one's own perception of boundaries in various situations. Spending more time with peers, outside of the family circle, children learn to build friendships.

\item {\bf Adolescent} (12-18 years old): 

This is a period of self-knowledge and identity forming, which sets off a hormonal explosion, often causes instability and unpredictability, poses challenges to educational performance and motivation. Adolescents attempt to discover how they can meaningfully contribute to society and how to present themselves based on their beliefs. This search pushes them away from the family and into the peer environment in preparation for the independent adult life, which becomes the leading activity at this stage. Dominant (abstract) logical thinking is no longer grounded in concrete cases and allows operating with abstract ideas and hypothetical concepts. This intellectual development encourages arguments and discussions in an increasingly mature manner. On the other hand, the perception of future is still limited to short-term outcomes, with significant importance placed on immediate gains and losses. This fixation on ``here and now'' passes by late adolescence.
\end{itemize}

\subsection{Age-appropriate activities}



Age provides key information about the tasks and activities which are natural and important from the developmental perspective at various stages, and therefore readily accepted by the child. Naturally engaging in such tasks, a child creates his or her own interpretation and interest, which stimulates the interaction in the long-term. This principle of matching content with age applies in any work with children -- be it educational, therapeutic, psychological, sports-related, social -- as well as in designing products for children (e.g. games, toys or books) \cite{copple2009developmentally, meschke2012developmentally, fisher2014designing, thibault2023handbook, kudrowitz2021designing, rocklin2001inside, ccer2016preparing}. In the following we provide concrete examples of age-appropriate\footnote{The term ``age-appropriate'' refers to a developmental concept whereby certain activities may be deemed appropriate or inappropriate to a child’s ``stage'' or level of development. Being ``age-appropriate'' means being suitable (in terms of topics, messages, and teaching methods) to the developmental and social maturity of the particular age or age group of children or adolescents, based on developing cognitive, emotional, and behavioral capacity typical for the age or age group \cite{UScode42sect710}.} activities for the age groups described in Sec.~\ref{sec:development-stages}. For an exhaustive guide on activities for all ages, we recommend \cite{harvard2014enhancing}.

\begin{itemize}
\item For an infant, age-appropriate activities could be games involving visual contact and emotional response, imitating simple actions, movements and sounds, manipulating objects, orienting in the direction of a stimulus, etc.
Participation of an adult is unavoidable, due to the central developmental task of building an emotional connection and trust.
\item For a toddler, age-appropriate activities could be manipulation and multi-modal contact with physical objects or navigation in space.
Peers or adults introducing new objects or ways to use them could sustain engagement.
\item For a preschooler, age-appropriate activities could be creative activities fostering the child to make something new or alter existing things, story-driven role-playing games, games involving memory, puzzles and other developmental elements.

\item For a gradschooler, age-appropriate activities should be purposeful in challenging and improving the logical capacity, for instance games which include rules and success criteria, feedback and progress communication. Personal achievements are important.

\item For an adolescent, developing a meaningful common interaction scenario is more challenging than targeting the interaction towards a specific child. This period varies greatly in its sub-stages, their duration and specific difficulties adolescents might face. An adolescent struggling with self-organization could benefit from study feedback and personalized tutoring.
Not having the patience to cope with a cumbersome or inefficient solution, a clear task or functional outcome of any activity designed for adolescents would be appreciated.
\end{itemize}

These tasks and occupations are already present in the child's daily activity, in kindergartens, schools or at home, in the form of toys and devices, board and computer games, designed to aid the child's development. A robot is well-equipped to be an embodied provider of these activities, acting as a toy or a training device, teacher or student, play companion or buddy as we described in Sec.~\ref{sec:robot-for-learning}.

\subsection{Age-appropriate CRI designs}

Evaluating existing CRI scenarios via the afore-presented lens of development appropriateness, we can identify several examples which account for age-specific requirements and therefore have the potential to stimulate meaningful long-term interactions.

The multipurpose robotic system for interactions with toddlers in \cite{vinoo2021design} is an example of an age-appropriate SAR that meets the activity requirement, providing interactions with physical objects.
The proposed system is a tele-operated child-sized robot, shaped as a steering wheel on a platform, which can initiate contact with children using non-verbal signals and engage them in an interaction (following, touching, pushing or pulling), where it acts both as a partner and a manipulable object (see Fig.~\ref{fig:cover}).
Moving around the free play space and engaging children into various activities allows for collecting new ideas about what else the robot could do
(e.g., holding a phone up to the robot’s ``ear'' and saying ``Hello'', using the robot as foot rest or cruising toy, or to augment toys by swirling play balls around the robot’s roll cage). 
This robot was primarily designed for early movement of children with disabilities and social engagement of young children during hospitalization, which bridges the long-term interaction design between healthcare and general-purpose CRI.

In the PepperRecycling educational game \cite{castellano2021pepperecycle}, which is intended to make children 7-9 y.o. more aware of and better disposed towards waste recycling,
two important conditions are met to stimulate the long-term engagement of gradschoolers. Firstly, the students are gaining new skills, and secondly, they do so in a game with rules and results, which provides feedback as a measure of increasing competence. The game is set as a competition between the robot Pepper and a child, which further stimulates competitive engagement and can ameliorate the learning gains. This competitive effect emerges also in other studies with gradschoolers, even when the interaction is not specifically designed for it (e.g. turn-taking trivia game with diabetes-related questions \cite{neggers2021investigating}, Snakes and Ladders game to promote vocabulary learning \cite{ahmad2019robot}).

Another scenario of educational play with Pepper is presented by Tanaka et al. \cite{tanaka2015pepper}, see Fig.~\ref{fig:cover}. Intended for preschoolers, the interaction relies less on competitive aspects and instead on role playing, which fits the target age group well. Authors created interactive content to learn English while having fun, designed to be used at home in small groups. In this scenario, children enacted a lesson where they solved tasks together with Pepper, given by an online tutor, and received rewards from the robot (``high five!'', a photo from Pepper's cameras, etc). This scenario minimizes the boring and mundane aspects of learning a language, instead exploiting the robot's physical embodiment and game-like features: students could switch the task, speak out the answers or press Pepper's buttons, bring and show an object to Pepper, etc. Role play with the robot can be a promising CRI direction for the preschooler target group.

The role play format was also used in a study by Okita et al. \cite{okita2011multimodal}, which aimed to investigate how different types of robot attention, nonverbal (e.g., nodding, looking) or verbal (e.g., “ok”, “uh huh”, “I see”), contribute to children’s affective behavior and prolonged engagement. The interaction (with children about 6 y.o.) was set as a cover story: the child's task is to prepare the robot to visit the zoo, where it would encounter two animals (a duck and a pig). Children were initially asked to share information they knew about the animals (e.g., the size, what they eat, how they communicate), and then told the robot a well-known story featuring those animals (i.e., “Ugly Duckling” or “Three Little Pigs and the Big Bad Wolf”). Talking about animals and storytelling are familiar activities for children of this age.
In this interaction scenario, the story part contributed to the active engagement of children over an extended period of time (two sessions, one per animal, each 35 minutes long, with a small break in between).

Another part of this study, which we already mentioned in Sec.~\ref{sec:robot-for-learning}, illustrates an important difference between preschoolers and gradschoolers engaged in the same game with the robot (table setting task). Younger children (4-5 y.o.) were more involved in the process, acting out their own vision of the game, which they found to be exciting in itself. In other words, what the robot said did not matter, as the game of response turn-taking was more important than its content. Differently, the older children (9-10 y.o.) aimed to grasp the content and adhere to the ``rules'' of the game, interactively reacting to the robot's actions and behavior.
This outcome is well aligned with the age-appropriate interaction structures described above.

Age-appropriate games for preschoolers should include freedom of creativity and flexible rule setting (i.e. the game itself defines the rules). In contrast to that, pre-defined rules and success criteria should be in place to engage gradschoolers into the game (i.e., the rules define the game)\cite{elkonin2005psychology}. Mismatching the game type with the age group will likely have an effect on the motivation and behavior of children during the game, and influence the outcome of the CRI.

For instance, Elgarf et al. \cite{elgarf2021once} test the verbal creativity of children working on storytelling with a creative (vs. non-creative) robot. The collaborative game is set as developing a fairy-tale story, in which the children and the robot produce and share their creative ideas to drive the plot forward. The story includes several characters (prince and princess) and creatures (crocodile, fish, bee, chicken, alien), which are shown on the screen and can be added to the ongoing story in an interactive window. The study is conducted with 9 y.o. participants (M=8.96 SD=0.58).
While the researchers expected the group working with the creative robot to produce more creative ideas, the experiment revealed no significant difference between the groups. In fact, the authors observed peculiar reactions to the robot in both conditions.
Some of the children did not like the interruption of the robot when it started proposing ideas. Some showed signs of frustration by telling the robot ``no'' each time it suggested an idea, others were irritated by the robot destroying their scenarios or did not like the interference of the robot with their story.
As outlined in Sec. \ref{sec:development-stages}, gradschoolers are interested in games with rules and in getting an objective evaluation of their own and others’ decisions. In this game, perhaps, they did not find the robot cooperative as it failed to develop their ideas, thus breaking the rules of the collaborative game. At the same time, the children might have seen the robot's own ideas as childish, illogical, and simply uninteresting, and did not like the story, eventually losing interest in interacting with the creative robot. It would be interesting to play this interaction with preschoolers, who engage in creative tasks in and of themselves more easily, and see if more positive responses and outcomes are elicited. Furthermore, the content of the game (a fairy-tale) could play out better for preschoolers.

When it comes to older children, it is worth highlighting that interactions structured as games could be less appropriate for adolescents, as their developmental tasks go beyond skill acquisition and competence evaluation (which is important for gradschoolers) and shift towards the social context. As such, a game with a robot could appear tiresome and unnecessary, which would render the interaction inefficient.
For instance, Alves-Oliveira et al. \cite{alves2019empathic} studied how an empathic robot behavior impacts the learning progress of 13-14 y.o. children in a collaborative Enercities game, which includes factual knowledge about different energy sources. This study reported the lack of learning gains.

Adolescents are more oriented towards social experience and interpersonal interactions than younger children and are more technically savvy to have less ``magical'' and more mature expectations from the robot.
If the interaction lacks social context, they might quickly lose interest in it, as noted by Serholt and Barendregt \cite{serholt2016robots}. Thirteen year old children, engaging in a learning game with a robot that could not respond to their attempts at social interaction, became notably less socially engaged (instead concentrating on the task performance). They went from being highly interactive with the robot to completely ignoring it. Conversely, younger children told the researcher that they still believed that the robot could understand them.

These examples showcase how the design of age-specific interaction scenarios has the potential for sustainable long-term engagement. Verifying this hypothesis in a series of studies, spanning diverse tasks and activities, is an appealing direction for future research.

\section{Individual characteristics}
\label{sec:individual-characteristics}

By introducing the periods of child development in Sec.~\ref{sec:development-stages}, we outlined the tasks and activities, which are common for children on various stages. Despite these commonalities, even similarly-aged children can and do behave differently in most situations, manifesting their individuality, which is based on several key psychological structures \cite{shiner2006temperament, thompson2010individual,shiner1998shall}:
\begin{itemize}
    \item \emph{Temperament} -- a cumulative set of the nervous system characteristics (e.g. its power, sensitivity, plasticity). Temperament defines, for instance, the ability of a child to work longer, or switch tasks quicker.
    \item \emph{Character} -- the typical emotional mood and vitality of the child. In their character children can be, for instance, cheerful, capricious, enthusiastic, passive, impulsive, thorough, confident, excitable, even, cautious as well as many others. Character traits can change throughout life depending on the specific development conditions, social environments and lived experiences.
    \item \emph{Personality} -- definitive individuality traits, consistent and stable, which reflect the child's perception of self and the environment, personal values and attitudes, preferred modes of interaction with others, as well as the means to reach the goals. Personality traits include, for instance, introversion and extraversion, rationality and sensitivity, conservatism and radicalism, assertiveness and agreeableness, confidence and anxiety, dominance and submission, egocentrism and altruism, among many others. The roots of these traits can be observed as early as preschool age, but they tend to fully form and stabilize following the self-awareness development occurring during adolescence.
    \item \emph{Aptitudes and  abilities}  -- internal presuppositions to succeed in different types of activities, e.g. cognitive, social, creative, educational, practical, etc.
    \item \emph{Interests, needs and preferences} -- individual motivational components, which drive and direct any activity, defining its relative importance and desirable outcome.
\end{itemize}
Illustrating the many aspects of individuality, these examples are just a short excerpt of how psychology describes children. For further reading, we recommend \cite{ eisenberg2006handbook,mroczek2014handbook,soto2015personality,stern2017psychology}.

CRI, as any other activity involving children such as education \cite{fontana1995psychology,snowman2014psychology}, inevitably faces the challenge of dealing with individual characteristics. Increased awareness thereof could thus significantly elevate the effectiveness and sustainability of child-robot interactions. 
For example, in Sec.~\ref{sec:robot-for-the-child} we noticed how individual characteristics may manifest themselves during the interaction and become a decisive factor for the outcomes of CRI studies \cite{kanda2004interactive,kanda2007two,michaud2005autonomous,konijn2020robot}. We also noticed how the lack of data on the participants' background may impede a proper assessment of how the robot's behavior influences the study outcomes \cite{baxter2017robot,yadollahi2018deictic}. 

In contrast to age, individual characteristics are difficult to recognize and formalize in the interaction scenario. To objectively measure them, experts in child development adopt quantitative and qualitative approaches, such as questionnaires, interviews, tests, observational methods etc. For further reading, we recommend \cite{greig2007doing,  whitcomb2013behavioral, sattler1988assessment}.

Our brief outline of the individuality structures aims to emphasise the complexity of inner psychological factors which create an individual portrait for each child. This complexity makes it most desirable that child development specialists participate in designing CRI scenarios to address the practical tasks in the fields of education, development, well-being as well as in the studies aiming to design new generations of social robots.

While the direct participation of child psychologists in the design and analysis of CRI studies is highly advised, it is not always possible.
In such cases, we encourage researchers to collect systematic background on all children prior to the experiment, in an interview or using a custom list of questions. This background data, as the previous sections highlight, can prove crucial to interpreting the experiment's results and improving follow-up interactions. If the experiment includes specific activities such as dancing exercises or puzzle solving, it is important to ask the participants about their prior experience of and attitude towards these tasks. Collecting children's responses about their interests (e.g., most and least favourite subjects at school, games, movies, free time activities), and their strengths and weaknesses (what do the teachers at school and parents at home often praise or otherwise) would be useful in a wide variety of scenarios, as much as it would be to understand the children's perception of and experience with robots in general. For the youngest participants, this information could be sourced from parents or kindergarteners. As we are talking about personal, often sensitive information about vulnerable participants, privacy and ethical reasons are imposed to reduce the amount of data collected and the number of people accessing it to the essential minimum. As such, careful consideration should be placed in the identification of the background information to be collected. Participatory research approaches, which aim to involve all relevant stakeholders in the research process \cite{maure2023participatory}, might also help to make parents, teachers, and even the children themselves better aware of the reasons why specific background information is asked for and how it will be used in the research and the interaction itself.

Furthermore, it is important to note the validity of the background checks. For instance, Fern\'andez-Llamas et al. \cite{fernandez2020analysing} report that the children could not correctly assess and report their knowledge level, which led to results lacking a significant impact of the robot. Similarly, Eimler et al. \cite{eimler2010following} note that the vocabulary proficiency of children in their study was based on information provided by the teacher, but in fact, the students could have learned the words elsewhere, which could have decreased the statistical significance of the study results.

Ultimately, the collected individual characteristics could (and should) be included in the CRI design with the aim of achieving a more enjoyable and efficient interaction. Such information, for instance, could be used by a robot to select an appropriate topic for a conversation or storytelling, assign roles in role-play and teamwork, define learning roles (e.g., teacher, peer, or tutee), and adapt its own emotional mode (e.g., level of neutrality). Following the child’s individual interests (for instance, drawing, singing, planes or astronomy), CRI scenario could include a set of joint activities for skill practicing or a deep-dive into the topic. Following the child’s individual challenges (for instance, improving memory as a basic skill for educational proficiency, or overcoming anxiety when presenting in front of classmates), it could offer appropriate exercises.

Explicitly emphasising this, Abe et al. \cite{abe2014toward} propose to design robotic playmates that can play with children accounting for their personalities. Having considered how shy children (aged 5-6 y.o.) engage in CRI, the authors point out that they may feel uneasy in the presence of a robot and propose a strategy to help them get accustomed to using play actions with a sense of security, such as quiz and hide-and-seek. On the contrary, other children, instead, after feeling a little bit nervous at the start, quickly get engaged in the interaction through the conversation with the robot.

Similar approaches to design interactions targeting a specific problem are adopted in the healthcare domain for children with special needs, which is key for rehabilitation and well-being 
e.g., to aid children with diabetes to learn how to manage their condition \cite{coninx2016towards, looije2008children}.
Further examples are covered in the recent reviews \cite{papakostas2021social,ismail2019leveraging}.
However, outside of studies focusing on children with special needs, only a handful of works match the interaction design with the target group and user needs. The difficulty stems from formulating a development target or identifying a problem to be addressed and selecting an appropriate interaction scenario to that end.

Equally important is to correctly define the CRI success criteria. If the robot's intervention aims to improve a specific skill or ability, success metrics should be validated instruments capable of capturing an objective difference in that skill or ability, measured pre- and post-experiment, as well as meaningful control groups, i.e., exercising without a robot. The number and duration of interactions play a key role in letting positive dynamics surface, as it is unrealistic to expect a significant and durable improvement in performance as a result of a single interaction, as short as 10 or 15 minutes \cite{ali2019can,elgarf2021once,gordon2015can}.

A skill-targeted approach is proposed in \cite{yadollahi2022motivating}, where the authors explore the potential of embodied activities with robots to aid the development of children’s spatial perspective-taking abilities. In this study, children 8-9 y.o. guided the robot along a maze by considering the robot’s point of view. To assess the effectiveness of the designed interaction, the authors repeated three tests to evaluate children’s perspective-taking abilities prior to and following the experiment. While the outcomes did not support the initial hypothesis that the robot intervention would improve spatial skills, they gave valuable insights into how the children perceive the environment through the robot's eyes. In particular, children seemed to follow one of the three strategies to guide the robot through the labyrinth: mentally rotating themselves or the robot, physically rotating themselves, or using trial and error to guide the robot. These differences likely arise from differences in the cognitive abilities and personality characteristics of the children -- different children will thus most likely perform differently on the task for reasons beyond the design of the robot and its behaviour.

A successful CRI scenario, oriented at a specific need of a child (improving poor handwriting), is presented in \cite{jacq2016building}. Assuming that poor self-confidence exacerbated and prevented resolving the writing difficulties, the authors designed a role-playing game in which the 5 y.o. child acted as a teacher for the robot. Four 1-hour-long sessions yielded positive outcomes, and the child was committed to the activity with the robot. Noteworthy, the game format of the interaction suited the age of the participant well.

A particularly significant effort to match the interaction scenario with age-specific needs is described by Michaud et al. \cite{michaud2005autonomous}. 
Starting with the Roball prototype (a self-propelling robotic ball that can sense its position and motion and thus the way it is being played with, which we discussed in Sec.~\ref{sec:robot-for-the-child}), the authors further adapted it for child development studies. 
The researchers started by seeking to determine the appropriate age of children to interact with Roball, and the age-specific needs that the robot can aim to fulfill.
Together with child development experts, they identified the target age group (6 to 24 months old children) and user needs (acquisition of general sensorimotor skills and mobility as a predominant factor in interplay situations). Matching the user needs with the Roball's properties confirmed the robot's utility for developing motor skills (crawling, walking), visual skills (for tracking an object or using visual effects), precise manipulations of the robot (to grasp Roball, make it spin, push it, or use the push buttons), intellectual skills (to explore trajectories and the effects of different actions while playing with Roball), social and emotional skills (by sharing the play with adults or other children). 
The design of Roball was developed targeting all these aspects to be part of the interaction. Roball was verified to be usable for interaction with children and maintaining their interest. 
The authors validated the robot in field studies and thoroughly reported how individual children interacted with the robot, though its actual impact on the long-term development goals remained out of scope in \cite{michaud2005autonomous}. 
The authors conclude that robotics and child development intersect in CRI, but not yet merge to accelerate cross-beneficial outcomes:
\vspace{2pt}
\begin{quote}
``Child development studies go far beyond the expertise of roboticists... However, people outside of robotics rarely have a realistic idea of how mobile robot be used. So, exposing robots such as Roball in a child development study is a discovery process and a demonstration of challenges and opportunities for both roboticists and child development experts, seeking and fostering strong collaborations between the two disciplines.'' \cite{michaud2005autonomous} (page 8)
\end{quote}
\vspace{2pt}
This consideration marks an important insight in the CRI domain. Robot interventions in a child's life indeed originate a discovery process in which robotics is put to use for specific developmental tasks. Nearly 20 years after their statement, we agree with and reprise the words of Michaud et al. to advocate for more prominence to be given to child-centered design. Input from child development studies could enrich the understanding of children's needs and help design and match the robot's capabilities to them. We argue that this would make interactions more efficient, sustainable, and enjoyable, thus advancing towards the ultimate goal of CRI.

\section{Conclusion and future work}

In this article, we investigate how children interact with robots.
Our central idea is that what happens during the interaction strongly depends on and therefore can be described by means of child psychology. 
Accordingly, in our review of CRI studies we aim to enrich the robot-centered perspective on designing robots to work with children, i.e. structuring the interaction around what the robot can technically do, with the child-centered perspective, i.e., structuring the interaction around the child's needs. We argue that such a multi-disciplinary perspective on CRI can help to identify the factors which influence children's behavior, design appropriate robot behaviours and interactions, select meaningful measurement criteria in the user studies, and correctly interpret the results and causalities in the observed behavior.

In line with the above motivation, in this article, we approach and review the development of the CRI scenarios through the optics of child development theories. This challenging task required gathering the bits of evidence scattered across the broad spectrum of CRI studies and structuring them according to child development frameworks, to propose a general picture that will hopefully help to plan, conduct, and evaluate CRI studies more efficiently.

In particular, we review the motivation and existing use-cases for deploying robots to work with children and outline the challenges that researchers face in describing the interaction results, often counter-intuitive or contradicting the prior art. We argue that the novelty effect, which often explains the positive affective outcomes and even immediate performance gains, complicates and hinders the scaling of existing designs to a long-term perspective. We analyze how children engage in the interaction process, and find and highlight evidence that individual differences significantly influence the outcomes and motivation to continue the interaction in the long-term. We note that the success of child-robot interaction depends on the effort that children put into it, and show how the format and content of the interaction, if and when aligned with age- and person-specific goals, can lead to natural and sustained engagement.

Given the variety of development stages and individual characteristics of children, we argue that the design of CRI experiments should be more informed by, or developed in collaboration with, child psychologists to ensure a balanced inclusion of the child-centered perspective. Ultimately, we believe that this would make the interactions more engaging and effective and their outcomes more conclusive.


In order to illustrate the child factor in the CRI studies, in this work
we focused mainly on those papers that report sufficient information on the individual children participating in the interaction with the robot, e.g. their behaviors, responses, differences in task performance, unusual and typical reactions. 
In the future, we intend to expand this work with a more systematic review of the state-of-the-art and the develop of a system of categories and keywords to describe CRI studies.

Having established the key role that individual characteristics play in the success of an interaction, and the importance of properly collecting and describing background facts to help separate the child factor from the robot factor, as future work we also plan to develop methods and procedures for the systematic collection of background information for children participating in CRI studies.

Looking beyond the current paradigm in which CRI scenarios are typically designed to address a single specific task, we note that the developmental process in children strongly involves experimentation, trial and error, overcoming challenges, and getting positive and negative reinforcement, in a very broad range of social, physical and intellectual circumstances. This observation is an interesting contrast to the idea that (adaptive) robots should be comfortable, enjoyable, and entertaining above all and in all circumstances. We envision general-purpose CRI robots that pose stimulating challenges to the children, attempting to activate their natural traits such as curiosity and creativity, cunning, stubbornness, and willfulness, and hope that this work will help motivate and support researchers in pursuing this direction.




%


\section*{Conflict of interest}
The authors declare that they have no conflict of interest.

\section*{Availability of data and material}
Not applicable

\section*{Code availability }
Not applicable
%



\bibliographystyle{IEEEtran}
\bibliography{bibliography}

\end{document}